%% file: StarVLA.tex
\newif\ifdvipreview
\ifdefined\directlua
  \ifnum\outputmode=0\relax
    \dvipreviewtrue
  \fi
\fi
\ifdvipreview
  \PassOptionsToPackage{dvipdfmx}{graphicx}
  \PassOptionsToPackage{dvipdfmx}{xcolor}
  \PassOptionsToPackage{dvipdfmx}{hyperref}
  \PassOptionsToPackage{dvipdfmx}{geometry}
\fi

\documentclass{article}

\usepackage{graphicx}
\usepackage{booktabs}
\usepackage{multirow}
\usepackage{algorithm}
\usepackage{algpseudocode}
\usepackage{amsmath}
\usepackage{tabularx}
\usepackage{textcomp}  
\usepackage{listings}

\usepackage[dvipsnames,table]{xcolor}
\usepackage{colortbl}
\usepackage{array}
\usepackage{rotating} 
\usepackage[numbers,compress]{natbib}

\definecolor{tableBest}{HTML}{B3CDE3}
\definecolor{tableSecond}{HTML}{E0ECF4}
\newcommand{\bestcell}[1]{\cellcolor{tableBest}\textbf{#1}}
\newcommand{\secondcell}[1]{\cellcolor{tableSecond}\underline{#1}}

\usepackage{graphicx} 
\usepackage{subcaption} 
\usepackage{caption}
\usepackage{enumitem}

\usepackage{pifont} 
\usepackage{xcolor} 
\usepackage{adjustbox} 

\usepackage{booktabs}
\usepackage{array}
\usepackage{adjustbox}
\usepackage{makecell}

\newcommand{\mypara}[1]{\smallskip\noindent\textbf{#1}}
\newcommand{\circnum}[1]{\textcircled{\scriptsize #1}}

\usepackage[final]{neurips_2024}
\reportdate{August 2026}
\reportprojectpage{\url{https://starvla.github.io/VLAct}}

\newtcolorbox{glancebox}{
  enhanced,
  colback=tableSecond!32!white,
  colframe=metablue!85!black,
  arc=4pt,
  boxrule=0.6pt,
  boxsep=2pt,
  left=7pt, right=7pt,
  top=5pt, bottom=4pt,
  drop shadow=gray!25!white
}
\newcommand{\glanceref}[1]{%
  \hfill\textcolor{metablue}{\textit{\S\ref*{#1}}}}
\newcommand{\glancetableref}[1]{%
  \hfill\textcolor{metablue}{\textit{Tab.~\ref*{#1}}}}

\ifdvipreview\else
\usepackage[utf8]{inputenc} 
\usepackage[T1]{fontenc}    
\fi
\usepackage{hyperref}       
\usepackage{url}            
\usepackage{booktabs}       
\ifdvipreview\else
\usepackage{amsfonts}       
\fi
\usepackage{nicefrac}       
\usepackage{microtype}      
\usepackage{xcolor}         

\usepackage{tocloft}

\usepackage{booktabs}
\usepackage{multirow}
\usepackage{graphicx}
\usepackage{xcolor}
\graphicspath{{RAW/}{RAW/figures/}{Figures/}}

\ifdvipreview
  \usepackage{fontspec}
  \usepackage{unicode-math}
  \defaultfontfeatures{
    Extension = .otf,
    RawFeature = {-no_designsize}}
  \newfontfamily\dvipreviewtitleface{texgyreadventor}[  
    UprightFont = *-regular, BoldFont       = *-bold,
    ItalicFont  = *-italic,  BoldItalicFont = *-bolditalic]
  \renewcommand{\titlefont}{\dvipreviewtitleface\bfseries}
  \DeclareRobustCommand{\circnum}[1]{%
    \leavevmode\hbox{%
      \setbox0\hbox{$\bigcirc$}%
      \rlap{\unhcopy0}\hbox to\wd0{\hss\raisebox{.3ex}{\tiny #1}\hss}}}
\fi

\input{RAW/title}

\begin{document}

\maketitle

\input{RAW/sections/0_abstract}

\tableofcontents
\clearpage  
\input{RAW/sections/1_introduction}

\input{RAW/sections/2_pilot}
\input{RAW/sections/3_method}
\input{RAW/sections/4_experiments}
\input{RAW/sections/5_analysis}
\input{RAW/sections/6_related_work}

\input{RAW/sections/7_conclusion}
\input{RAW/sections/Authors}

\clearpage


{\small

\bibliographystyle{plainnat}  
\bibliography{main}
}

\clearpage
\clearpage
\input{RAW/sections/X_appendix}
\clearpage






\end{document}

%% file: RAW/title.tex
\title{\fontsize{14.5pt}{18pt}\selectfont
Beyond Data Scaling: Representation-Centric\\
Continued Pre-training for Vision-Language-Action Models}
\author{}

%% file: RAW/sections/0_abstract.tex
\begin{abstract}
Scaling robot data is crucial for building generalist Vision-Language-Action (VLA) models, yet robot trajectories are fundamentally harder to scale than web-scale image-text data because they require embodied collection and sparsely cover the physical world. This makes representation quality a central bottleneck: under a fixed robot-data budget, VLA continued pre-training must convert limited trajectories into transferable visual-action knowledge, rather than merely fit actions.
We propose \textbf{VLAct}, a VLA-oriented VLM backbone trained with a representation-centric continued pre-training recipe, which starts from a pretrained VLM and trains on broad, heterogeneous, multi-embodiment robot data before downstream task-specific fine-tuning. VLAct preserves the broad VLM prior, avoids over-specializing the backbone to a single action head, and encourages shared action semantics across embodiments through VLM-prior preservation, multi-head continuous action co-supervision, and a partially unified cross-embodiment action layout, while leaving downstream users free to attach task-specific action heads during fine-tuning.
Across multi-embodiment simulation benchmarks, real-world robot experiments, and unseen-embodiment transfer, VLAct consistently improves downstream performance under fixed fine-tuning protocols. On LIBERO-Plus and RoboTwin~2.0, VLAct surpasses large-scale industrial VLA systems such as ABot-M0 and LingBot-VLA, achieving 82.6\% and 92.5\% success, respectively. On the recently published RoboDojo simulation benchmark, VLAct achieves a 10.66 average score and 7.60\% success rate, ranking sixth among all policies by success rate. It outperforms all explicitly designated world-action-model (WAM) entries on both metrics, as well as several industry-developed systems. Most notably, on RoboCasa-GR1, a humanoid embodiment never seen during continued pre-training, \textbf{VLAct with only 20\% of downstream trajectories already outperforms the full-data GR00T-N1.6 baseline.} These results are obtained using fully open-source data and only a \textbf{16-GPU} training setup, showing that representation-centric continued pre-training can deliver highly competitive performance under a modest compute budget and is an important independent axis of VLA progress beyond data scaling.
All models and training pipelines are open-sourced.
  
\end{abstract}

\begin{glancebox}
{\noindent\sffamily\bfseries Highlights at a Glance}\par
{\color{metablue}\hrule height 0.35pt}\vspace{3pt}
\begin{enumerate}[
  leftmargin=2.2em,
  itemsep=3pt,
  parsep=0pt,
  topsep=1pt,
  label={\small\sffamily\bfseries\textcolor{metablue}{H\arabic*.}}
]
  \item \textbf{Foundation VLMs for physical tasks.} Beyond a specific policy, VLAct studies how to make foundation VLMs better suited to physical tasks and stronger foundations for VLA. \glanceref{sec:intro}
  \item \textbf{Representation-centric continued pre-training.} VLAct preserves VLM priors, diversifies action-head supervision, and shares action semantics across embodiments. \glanceref{sec:method}
  \item \textbf{Strong across benchmarks.} VLAct reaches 82.6\% on LIBERO-Plus, 54.8\% on VLA-Arena, and 92.5\% on RoboTwin~2.0. \glanceref{sec:experiments}
  \item \textbf{Data-efficient embodiment transfer.} With only 20\% of RoboCasa-GR1 trajectories, VLAct reaches 49.5\%, exceeding full-data GR00T-N1.6 at 47.6\%. \glanceref{sec:exp-cross}
  \item \textbf{Open and efficient scaling.} With open data and 16 GPUs, VLAct ranks 6th of 35 policies by RoboDojo success and beats every designated WAM on both metrics. \glancetableref{tab:robodojo}
\end{enumerate}
\end{glancebox}

\clearpage

%% file: RAW/sections/1_introduction.tex
\section{Introduction}
\label{sec:intro}

\begin{figure}[H]
\centering
\includegraphics[width=\linewidth]{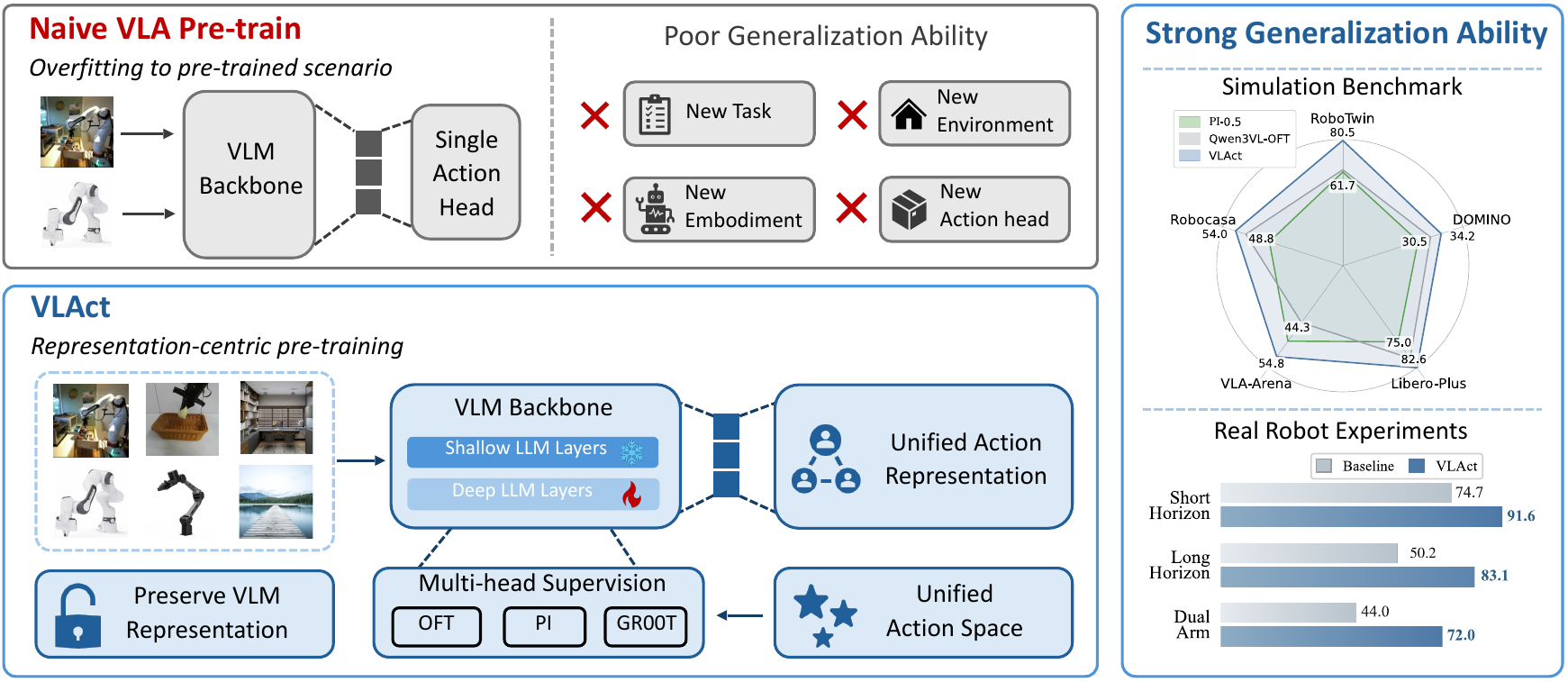}
\vspace{-0.4cm}
\caption{\textbf{Overview of VLAct.} VLAct adopts representation-centric continued pre-training to learn generalizable action-aware representations across tasks, embodiments, environments, and action heads.}
\label{fig:teaser}
\end{figure}

Scaling data, model size, and training compute has been a central driver behind modern foundation models~\cite{achiam2023gpt,team2023gemini}. In language and vision, this scaling paradigm succeeds not simply because web-scale corpora are large~\cite{liu2024improved,dai2023instructblip,bai2025qwen3}, but because they provide broad coverage of visual and semantic variation. Vision-Language-Action (VLA) models naturally aspire to the same paradigm: more robot trajectories should, in principle, support more generalist robot policies~\cite{black2024pi0, zitkovich2023rt, bjorck2025gr00t, kim2024openvla, yang2026abot}. Yet robot learning faces a qualitatively different scaling bottleneck.

Robot trajectories cannot be scraped from the web. They must be produced through embodied execution in the physical world, often under human teleoperation or carefully designed data-collection protocols\cite{o2024open,khazatsky2024droid,wu2025robocoin,lee2025molmoact}. More importantly, the space that robot policies must generalize over is combinatorial and continuous, spanning scenes, objects, task goals, embodiments, and contact-rich dynamics. As a result, even very large robot datasets remain sparse samples of the physical interaction space, with partial and uneven coverage across the situations a robot may encounter.

This does not diminish the value of scale, but it changes the role scale should play. If VLA continued pre-training cannot count on exhaustive coverage~\cite{o2024open,zitkovich2023rt}, then its success depends not only on how many trajectories are collected, but on how effectively those trajectories induce transferable visual-action representations. We therefore look \emph{beyond data scaling}: not to replace scale, but to ask what must accompany it. \textbf{\emph{Given a fixed robot-data budget, how can VLA continued pre-training learn strong representations that generalize beyond the trajectories on which they were trained?}}

This paper takes a representation-centric view of VLA continued pre-training. Rather than treating robot continued pre-training merely as large-scale action fitting, we view it as a process of distilling robot trajectories into reusable visual-action knowledge within the backbone. A useful VLA backbone should not only predict the actions observed in the pre-training data, but also internalize transferable priors about objects, affordances, spatial relations, and action-conditioned physical interactions that downstream policies can reuse across new tasks, scenes, and embodiments~\cite{huang2023voxposer,driess2023palm}. Effective VLA continued pre-training therefore requires designing how trajectory data shapes the backbone representation, rather than simply optimizing action prediction on the pre-training distribution.

\begin{figure}[t]
    \centering
    \vspace{-0.2cm}
    \includegraphics[width=1\linewidth]{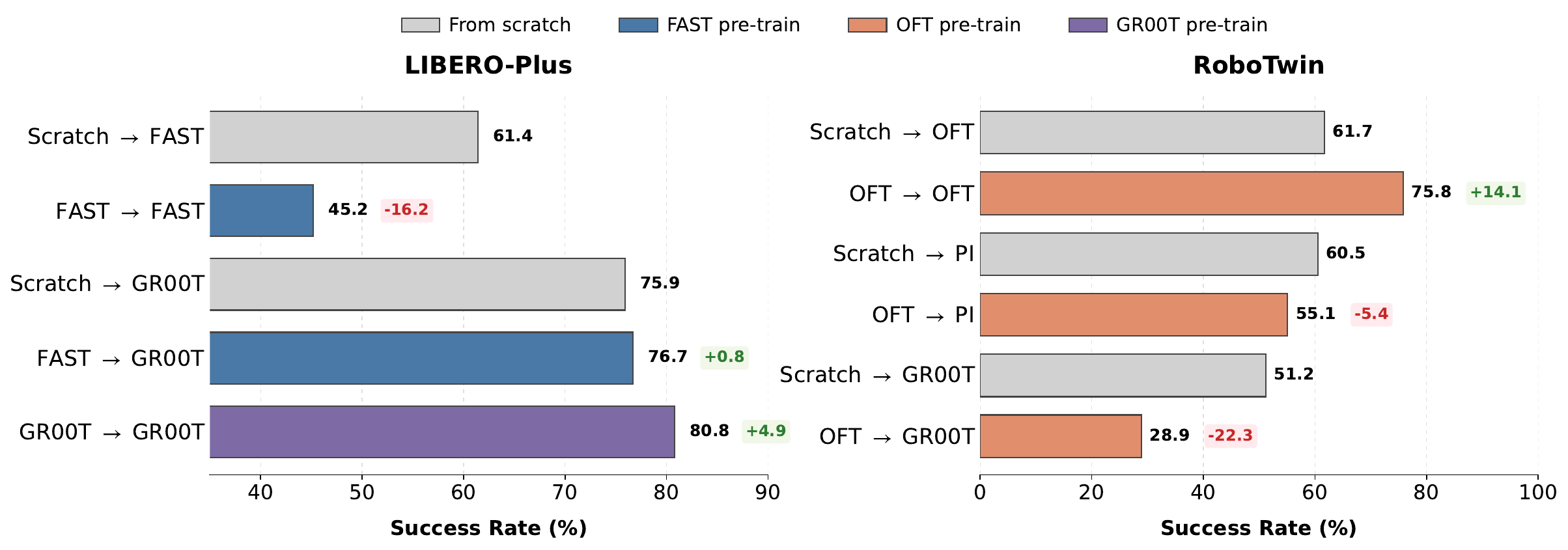}
    \vspace{-0.4cm}
\caption{\textbf{Action supervision reshapes VLA backbone representations.}
Keeping the VLM backbone fixed, we vary the action head used during pre-training and fine-tuning. Discrete FAST supervision can transfer to a continuous GR00T head, but retaining the FAST head suffers from discretization-induced information loss. Continuous OFT supervision improves same-head fine-tuning, but transfers poorly to other continuous heads, suggesting head-specific representation collapse rather than a generally reusable action representation.}
\vspace{-0.4cm}
\label{fig:naive-fails}
\end{figure}

To build a transferable VLA backbone, we first examine how robot continued pre-training shapes the learned representation, rather than only its action-prediction accuracy. We identify three failure modes in naive VLA continued pre-training. First, because robot trajectories are much narrower than the web-scale corpora used to train VLMs, full end-to-end updating can erode broadly useful vision-language features. Second, supervision from a single action head can over-specialize the backbone to that head's decoding geometry, limiting transfer to alternative heads. Third, embodiment-specific output spaces can weaken cross-embodiment sharing by representing physically comparable actions, such as gripper opening and closing, through isolated robot-specific heads.

Motivated by these observations, we propose \textbf{VLAct}, a VLA-oriented VLM backbone trained with a representation-centric continued pre-training recipe. VLAct preserves the VLM prior through shallow-layer protection and caption-data mixing, prevents single-head over-specialization through multi-head continuous action co-supervision, and encourages shared action semantics through a partially unified cross-embodiment action layout with a wrap-aware loss for periodic joints. These components are used only during continued pre-training to shape the backbone representation, while downstream users remain free to attach the action head best suited to their task and fine-tune under their own data budget.

\mypara{What ``continued pre-training'' means here.}
We start from an \emph{already pretrained} VLM and train it on broad, multi-embodiment robot trajectories before any downstream task-specific fine-tuning, rather than pre-training a foundation model from scratch. Representative generalist VLA systems, including $\pi_0$~\cite{black2024pi0}, $\pi_{0.5}$~\cite{intelligence2025pi_}, and GR00T~N1/N1.5~\cite{bjorck2025gr00t}, adopt exactly this setting and commonly call it \emph{VLA pre-training}; we adopt the more precise term here, and keep ``pre-training'' as shorthand where the stage is clear from context.

We evaluate \textbf{VLAct} across multi-embodiment simulation benchmarks, real-world robot experiments, and unseen-embodiment transfer. Under a fixed downstream protocol, it consistently improves over generic VLM backbones and naive VLA-pretrained baselines.
VLAct outperforms strong public and industrial VLA systems across benchmarks: 82.6\% on LIBERO-Plus~\cite{fei2025libero}, 54.8\% on VLA-Arena~\cite{zhang2025vla}, and 92.5\% on RoboTwin~2.0~\cite{chen2025robotwin}. On recent RoboDojo~\cite{chen2026robodojo}, it achieves a 10.66 score and 7.60\% success rate, ranking sixth by success and outperforming all explicitly designated WAM models. With only 20\% of RoboCasa-GR1 data, it also surpasses the full-data GR00T-N1.6 baseline~\cite{bjorck2025gr00t}. All these results are obtained using fully open-source data and only a \textbf{16-GPU} training setup.

Furthermore, across every VLAct results comparison, \textbf{the VLM backbone weights are the only thing that changes}: the action head and its fresh initialization, the downstream data, the optimizer, and the fine-tuning budget are all identical. The resulting \textbf{7.6--21.4} point gains are therefore attributable to the backbone alone.
Hence, we take this as evidence that the VLM should be treated not as a fixed component inherited from general vision-language pre-training, but as a first-order design variable for VLA. We hope this encourages the community to study \textbf{how foundation VLMs can be made better suited to physical tasks and serve as stronger foundations for VLA models.}

%% file: RAW/sections/2_pilot.tex
\section{Pilot Study: How Action Supervision Shapes VLA Backbones}
\label{sec:pilot}

A central goal of VLA continued pre-training is to produce a reusable backbone, rather than a policy initialization tied to one particular action head. In practice, there is no universally optimal action head: discrete token heads (FAST), regression heads (OFT), flow-matching heads (PI), and diffusion-style continuous heads (GR00T) impose different inductive biases and may be preferred under different tasks, embodiments, horizons, or deployment constraints. A useful pretrained backbone should therefore expose action-relevant information in a form that downstream heads can reuse, instead of encoding it in a representation specialized to the head used during pre-training.

As shown in Fig.~\ref{fig:naive-fails}, we isolate this action-head axis through controlled pilot experiments on LIBERO-Plus and the RoboTwin-Clean setting. Fixing the VLM backbone to Qwen3-VL-4B, we vary the action head used during pre-training and fine-tuning, and test how different heads reshape the resulting backbone. We use representative discrete and continuous heads, and the head definitions are shown in Appendix~\ref{sec:head_detail_raw}.

\mypara{Discrete supervision can transfer, but suffers from discretization-induced information loss.}
The LIBERO-Plus probe asks whether discrete FAST-token pre-training provides sufficient action information for downstream control. FAST supervision can inject transferable structure into the backbone: when the FAST-pretrained backbone is paired with a continuous GR00T head, performance slightly improves over from-scratch GR00T fine-tuning. However, retaining the FAST Action Head performs much worse than continuous-head fine-tuning, and FAST pre-training does not close this gap. This suggests that discrete action tokens teach coarse action structure, but lose fine-grained temporal and amplitude information that matters for manipulation.

\mypara{Single-head continuous supervision can induce head-specific representation collapse.}
The RoboTwin-Clean probe shows a different failure mode. OFT pre-training substantially improves downstream performance when the fine-tuning head is also OFT, confirming that continuous action supervision can inject fine-grained action information into the backbone. However, this gain does not transfer to other continuous heads: the same OFT-pretrained backbone performs poorly when paired with PI or GR00T. This suggests that the learned representation is not simply more action-aware in a head-agnostic sense. Instead, single-head pre-training can collapse the backbone feature geometry toward the directions that are most useful for the pre-training head. Action-relevant information may remain present, but it is organized in a head-specific form that other downstream heads cannot easily decode. Strong same-head performance therefore overstates backbone reusability.


\mypara{Analysis.}
Action supervision is not neutral: the pre-training head directly shapes the backbone representation. Discrete supervision can transfer across heads, but loses fine-grained action information through discretization. Continuous supervision avoids this bottleneck, but single-head pre-training can cause head-specific representation collapse, organizing action features around one head geometry rather than keeping them broadly accessible. Thus, a transferable VLA backbone should preserve fine-grained action information while supporting multiple downstream heads. This motivates VLAct's multi-head continuous co-supervision in Sec.~\ref{sec:method-align}.

%% file: RAW/sections/3_method.tex
\section{VLAct: Representation-Centric VLA Continued Pre-training}
\label{sec:method}

\begin{figure}
    \centering
    \includegraphics[width=1\linewidth]{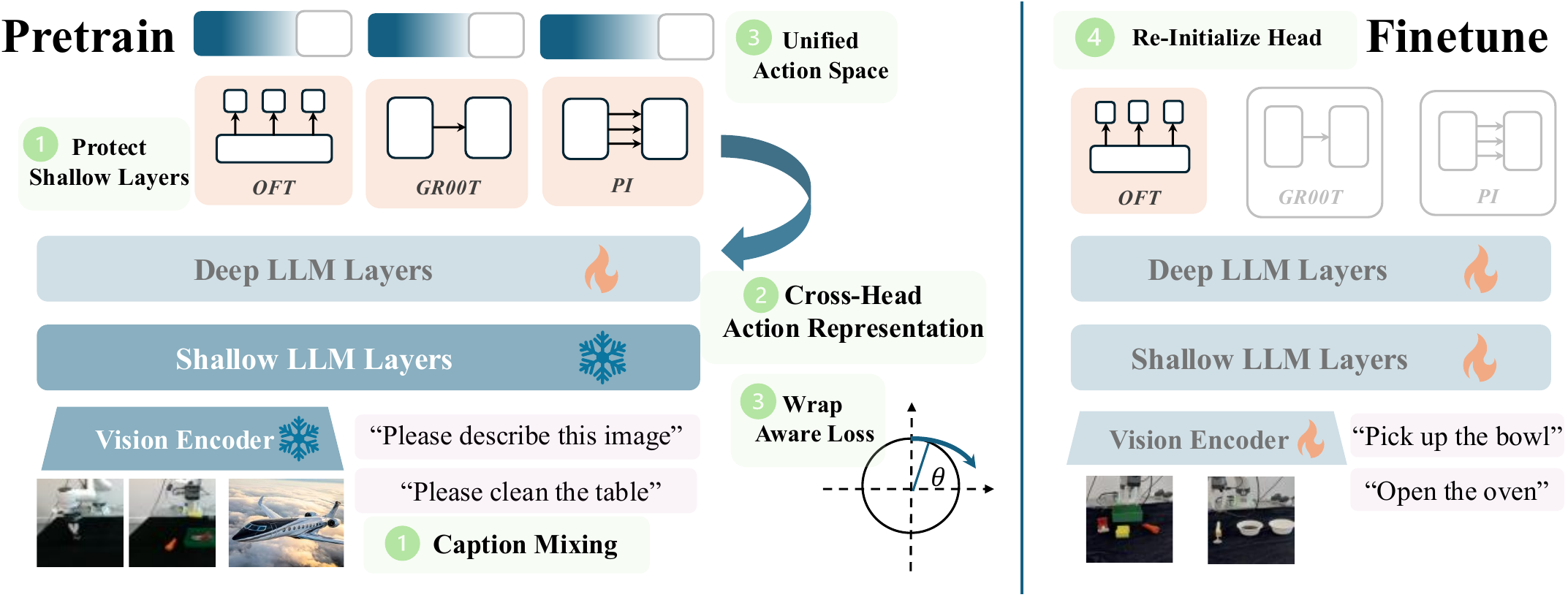}
    \caption{\textbf{Overview of VLAct.}
During pre-training, \circnum{1} VLAct preserves the VLM prior through shallow-layer protection and caption mixing (\S~\ref{sec:method-preserve}), \circnum{2} uses multi-head continuous supervision to avoid head-specific representation collapse (\S~\ref{sec:method-align}), and \circnum{3} aligns cross-embodiment action semantics with a unified action layout and wrap-aware loss (\S~\ref{sec:method-unify}). During fine-tuning, \circnum{4} we reuse the pretrained backbone and freshly initialize the task-specific action head (\S~\ref{sec:method-recipe}).}
\vspace{-0.5cm}
    \label{fig:method-overview}
\end{figure}

\subsection{Overall Continued Pre-training and Fine-tuning Recipe}
\label{sec:method-recipe}

As shown in Fig.~\ref{fig:method-overview}, VLAct performs continued pre-training of a reusable VLA backbone from a VLM initialization. During continued pre-training, it preserves the VLM prior by freezing the vision encoder and shallow LLM layers, while mixing robot trajectories with caption data. Robot samples supervise the shared backbone through multiple continuous action heads, including OFT, PI, and GR00T, under a partially unified action layout with masked inactive dimensions and wrap-aware losses.

These continued pre-training components are used only to shape the backbone. During fine-tuning, we discard the pre-training heads and caption stream, attach a freshly initialized task-specific action head, and train under the same downstream setting as each baseline. Thus, improvements come from the learned backbone representation rather than from reusing a pre-adapted action head.

\subsection{Preserving the VLM's Vision-Language Representation}
\label{sec:method-preserve}

\mypara{Motivation.}
A VLA backbone starts from a strong vision-language model, such as Qwen3-VL, whose visual and language representations are learned from broad web-scale data. Naive action-driven continued pre-training can overwrite this prior: robotics trajectories are expensive, narrow in visual diversity, and concentrated on a limited set of manipulation scenes. As a result, end-to-end VLA continued pre-training may improve action prediction on the pre-training distribution while degrading the general vision-language features that make the backbone transferable. Our first design goal is therefore to inject action knowledge without unnecessarily drifting away from the original VLM representation. We use two simple mechanisms: shallow-layer protection and caption-mixed pre-training.

\mypara{Shallow-layer protection.}
We freeze the entire vision encoder and the lower half of the LLM layers during VLA pre-training, while updating the upper LLM layers and the action heads. The motivation is to protect the parts of the model most responsible for low-level visual processing and early vision-language alignment, while still allowing the upper layers to adapt to action-conditioned reasoning. During downstream fine-tuning, we unfreeze the full model so that all parameters can adapt to the target task. This simple freezing strategy improves downstream performance by 3.7\% on LIBERO-Plus and 3.4\% on Agilex. Detailed ablations are provided in Appendix~\ref{sec:freeze_detail}.

\mypara{Caption-mixed pre-training.}
Mixing VLM data during VLA pre-training is a common strategy for mitigating representation drift, but not all VLM supervision is equally useful. We ablate several VLM data mixtures and find that captioning data provides the strongest anchor for downstream performance. The reason is intuitive: captions offer dense supervision over objects, attributes, spatial relations, and scene context, helping preserve the backbone's original vision-language representation while it learns from robot trajectories. We therefore use caption-mixed pre-training in VLAct. Detailed ablations are provided in Appendix~\ref{sec:vlm_data}.

\mypara{Takeaway~1.}
\emph{VLAct preserves the VLM prior while injecting action knowledge during continued pre-training: shallow-layer protection limits representation drift, and caption mixing anchors the trainable layers with dense vision-language supervision.}



\subsection{Aligning and Diversifying the Action Representation}
\label{sec:method-align}

\mypara{Design principle.}
The pilot study in Sec.~\ref{sec:pilot} suggests that an effective VLA backbone should satisfy two requirements. First, it should support \emph{head transfer}: downstream users should be able to attach action heads suited to their target embodiments and deployment scenarios, without inheriting a representation specialized to one particular head geometry. Second, it should learn high-quality action features, rather than merely fitting the pre-training head. Naive single-head pre-training can violate these requirements. Since the backbone and the attached head are optimized jointly, the backbone may encode action information in a form tailored to that head. This head-specific feature specialization reduces transferability across heads, even if the resulting representation performs well when the same head is reused.

\mypara{Co-supervised multi-head pre-training.}
We address this by pre-training the same backbone with multiple action heads in parallel. Concretely, we attach three representative continuous heads, \emph{OFT}, \emph{PI}, and \emph{GR00T}, to the shared VLA backbone. Given the same vision-language input, the backbone produces a shared latent representation $z$. Each head receives the same $z$ and predicts the same ground-truth action chunk $a$. The training objective is
\[
    \mathcal{L}_{\mathrm{action}}
    =
    \mathcal{L}_{\mathrm{OFT}}
    +
    \mathcal{L}_{\mathrm{PI}}
    +
    \mathcal{L}_{\mathrm{GR00T}} .
\]
This objective prevents the representation from becoming tied to a single decoder: the backbone must expose action information in a form that can be used by several action parameterizations. Since all heads share the same backbone forward pass, multi-head supervision adds only lightweight head-specific computation rather than repeated backbone computation.

This design is deliberately simple: we do not introduce a new action head or a specialized alignment module. Instead, we use head diversity itself as supervision. Because the three heads impose different objectives and decoder biases on the same action-prediction problem, the backbone cannot rely on features that are useful only to one head. To reduce all losses simultaneously, it must encode action information in a form that remains accessible across multiple downstream parameterizations.

Co-supervision therefore plays two roles. It makes the backbone more head-agnostic, improving transfer to alternative downstream heads; and it regularizes the representation, yielding stronger action features even when the downstream head matches one of the pre-training heads. As shown in Appendix~\ref{sec:appendix_heads}, this second effect is empirically important: multi-head pre-training not only improves head transfer, but can also outperform single-head pre-training under the same downstream head.

\mypara{Takeaway~2.}
\emph{Multi-head co-supervision turns head diversity into a representation-learning signal. By training OFT, PI, and GR00T on the same backbone latent and action targets, VLAct forces action information to remain accessible across different head geometries rather than collapsing toward one of them. This produces a backbone that transfers better across downstream heads and learns stronger action features than single-head pre-training.}

\subsection{Unifying the Action Representation across Embodiments}
\label{sec:method-unify}

\begin{figure}[t]
    \centering
    \includegraphics[width=1.\linewidth]{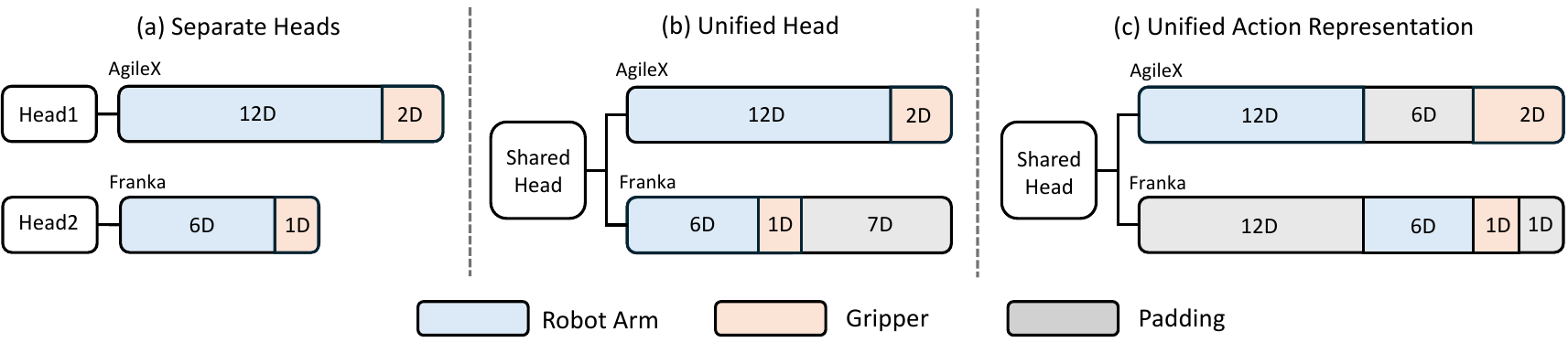}
    \vspace{-0.5cm}
\caption{\textbf{Unified action space design.}
We compare separate embodiment-specific heads, a naively unified action space, and our partially unified action space. VLAct shares physically aligned dimensions, such as gripper open/close, while masking unavailable or embodiment-specific dimensions.}
    \vspace{-0.6cm}

    \label{fig:unified-head}
\end{figure}

\mypara{Motivation.}
The previous Sec.~\ref{sec:method-align} addresses variation across action heads. We now consider variation across robot embodiments. Different embodiments have different action spaces, and a common solution is to attach a separate action head or output projector to each robot type. This is flexible, but it also hides any shared action structure inside embodiment-specific heads. Our motivation is simple: when two embodiments share a physically meaningful action dimension, the supervision for that dimension should also be shared; when they do not, we should not force an artificial alignment.

\mypara{Partially unified cross-embodiment action space.}
To share action supervision across embodiments without forcing incompatible
degrees of freedom into the same coordinates, VLAct uses a partially unified
cross-embodiment action space. Fig.~\ref{fig:unified-head} compares three
natural designs. The first uses embodiment-specific heads, which isolate
supervision across robots. The second uses a fully unified action space, where
lower-dimensional robots are padded to a common action dimension; this can
naively align coordinates with different physical meanings. VLAct adopts the
third design: one shared action head with an action space that is unified only
along physically comparable dimensions.

Concretely, gripper dimensions are shared across embodiments because
open/close commands have comparable semantics, while arm dimensions remain
embodiment-specific when the robots have different kinematics or degrees of
freedom. During training, each sample contributes loss only on the active
dimensions for its embodiment, and inactive dimensions are masked out. This
keeps the architecture minimal: no embodiment adapter, routing module, or
embodiment-conditioned decoder is introduced. Instead, cross-embodiment sharing
is induced directly through shared coordinates in the action space. As shown in
Appendix~\ref{sec:unified-output-layout}, this partial unification improves
downstream performance by sharing supervision where action semantics are
physically aligned while avoiding naive full sharing across incompatible joints.

\mypara{Wrap-aware loss for periodic joint angles.}
We also handle a small but important parameterization issue for absolute joint
angles. Standard regression treats $179^\circ$ and $-179^\circ$ as
$358^\circ$ apart, although they are physically only $2^\circ$ apart. We
therefore apply a wrap-aware loss on periodic joint-angle dimensions, measuring
angular residuals modulo $360^\circ$. This correction is only used for absolute
joint dimensions; the full definition and dimension scope are provided in
Appendix~\ref{sec:unified-output-layout}.

\mypara{Takeaway~3.}
\emph{Cross-embodiment continued pre-training benefits from sharing supervision where
semantics are physically aligned. VLAct uses one shared head with
a partially unified action space: gripper dimensions are shared across
embodiments, while incompatible arm dimensions remain embodiment-specific and
masked when inactive. This avoids isolated embodiment-specific heads and naive
full sharing, yielding stronger cross-embodiment representations and better
downstream performance.}

%% file: RAW/sections/4_experiments.tex
\section{Experiments}
\label{sec:experiments}
VLAct is pre-trained on fully open-source robot datasets, including DROID~\citep{khazatsky2024droid}, InternA1~\citep{tian2025interndata}, RoboCoin \citep{wu2025robocoin}, and MolmoAct \citep{lee2025molmoact}, together with captioning data for VLM representation preservation. Data processing and cleaning details are provided in Appendix~\ref{sec:data_clean_detail}. We build on the StarVLA training codebase ~\citep{community2026starvla} and use Qwen3-VL-4B as the base vision-language backbone. All experiments are conducted on 16 GPUs. We will release the training scripts and model checkpoints.

\subsection{Effectiveness on Single-arm Franka Manipulation}
\label{sec:exp-franka}

\mypara{Evaluation benchmarks.}
We evaluate single-arm Franka manipulation mainly on \textit{LIBERO-Plus}~\citep{fei2025libero}, a recent benchmark designed to probe robustness beyond the saturated standard LIBERO setting. LIBERO-Plus introduces systematic test-time perturbations, including changes in camera views, robot states, visual noise, object layouts, and task instructions, making it suitable for evaluating transferable VLA continued pre-training. We compare VLAct with representative VLA systems, including $\pi_0$~\citep{black2024pi0}, $\pi_{0.5}$~\citep{intelligence2025pi_}, and Abot-M0~\citep{yang2026abot}. Baseline numbers are taken from LIBERO-Plus~\citep{fei2025libero}. Additional Franka results, VLA-Arena, are reported in Appendix~\ref{sec:add_franka}.

\mypara{LIBERO-Plus results.} 
Table~\ref{tab:libero-plus} shows that our VLAct achieves the best overall score, reaching \textbf{82.6\%}, and improves over our strongest in-house baseline, Qwen3VL-OFT, by \textbf{7.6\%} (82.6\% \emph{vs.} 75.0\%). This comparison isolates the effect of the continued pre-training recipe, since both models use the same Qwen3-VL backbone family and an OFT downstream head. VLAct also surpasses Abot-M0, a large-scale VLA system from Alibaba, by 2.1\%, using fully open-source data and substantially fewer training resources. The largest gains over Qwen3VL-OFT appear on \emph{Camera}, \emph{Robot}, \emph{Noise}, and \emph{Layout}, suggesting that representation-centric continued pre-training yields more robust visual-spatial representations than standard Qwen3-VL fine-tuning on LIBERO.

\begin{table}[t]
\centering
\caption{\textbf{Results on LIBERO-Plus.} Per-dimension success rate (\%) on the LIBERO-Plus robustness benchmark. All models are trained on standard LIBERO and evaluated on the perturbed test set. Dark blue highlights VLAct and per-column best results; light blue highlights the strongest prior method by total score and per-column second-best results.}
\label{tab:libero-plus}
\begin{tabular}{lcccccccc}
\toprule
\textbf{Method} & \textbf{Camera} & \textbf{Robot} & \textbf{Lang.} & \textbf{Light} & \textbf{Bg.} & \textbf{Noise} & \textbf{Layout} & \textbf{Total} \\
\midrule
OpenVLA~\citep{kim2024openvla}        &  0.8 &  3.5 & 23.0 &  8.1 & 34.8 & 15.2 & 28.5 & 15.6 \\
OpenVLA-OFT~\citep{kim2025fine}        & 56.4 & 31.9 & 79.5 & 88.7 & 93.3 & 75.8 & 74.2 & 69.6 \\
NORA~\citep{hung2025nora}             &  2.2 & 37.0 & 65.1 & 45.7 & 58.6 & 12.8 & 62.1 & 39.0 \\
WorldVLA~\citep{cen2025worldvla}      &  0.1 & 27.9 & 41.6 & 43.7 & 17.1 & 10.9 & 38.0 & 25.0 \\
UniVLA~\citep{wang2025unified}           &  1.8 & 46.2 & 69.6 & 69.0 & 81.0 & 21.2 & 31.9 & 42.9 \\
$\pi_0$~\citep{black2024pi0}          & 13.8 &  6.0 & 58.8 & 85.0 & 81.4 & 79.0 & 68.8 & 53.6 \\
$\pi_0$-FAST~\citep{pertsch2025fast}  & \secondcell{65.1} & 21.6 & 61.0 & 73.2 & 73.2 & 74.4 & 68.8 & 61.6 \\
RIPT-VLA~\citep{tan2025interactive}          & 55.2 & 31.2 & 77.5 & 88.3 & 91.6 & 73.5 & 74.2 & 68.4 \\
\rowcolor{tableSecond}
Abot-M0~\citep{yang2026abot}     & 60.4 & \secondcell{67.9} & \secondcell{86.4} & 96.2 & 91.6 & \bestcell{86.4} & \secondcell{82.6} & \secondcell{80.5} \\
Qwen3VL-OFT                            & 47.0 & 60.1 & \bestcell{87.0} & \secondcell{96.3} & \secondcell{95.3} & 73.1 & 79.2 & 75.0 \\
\midrule
\rowcolor{tableBest}
\textbf{VLAct}                         & \bestcell{73.9} & \bestcell{68.4} & 81.5 & \bestcell{96.7} & \bestcell{96.7} & \secondcell{86.0} & \bestcell{83.3} & \bestcell{82.6} \\
\bottomrule
\end{tabular}

\end{table}

\subsection{Effectiveness on Bimanual AgileX Manipulation}
\label{sec:exp-agilex}

\mypara{Evaluation benchmarks.}
To evaluate bimanual AgileX manipulation, we use two recent benchmarks with complementary generalization axes: \textit{RoboTwin~2.0}~\citep{chen2025robotwin} and \textit{DOMINO}~\citep{fang2026towards}. RoboTwin~2.0 evaluates robust dual-arm manipulation under clean scenes and randomized visual conditions. We report two settings: the \emph{Base} setting uses only 50 clean trajectories per task, while the \emph{Data Scaling} setting further adds 500 domain-randomized expert trajectories per task. The latter are successful expert demonstrations released by the RoboTwin~2.0 team under its official \texttt{demo\_randomized} setting, in which background, clutter, tabletop height, lighting, and other scene factors are varied; they are \emph{not} random-action rollouts. Unless otherwise noted in the cited paper, Data Scaling comparisons follow this multi-task protocol with 2,500 clean and 25,000 randomized demonstrations. Training compute, optimization schedules, and checkpoint-selection procedures may nevertheless differ across methods.
This allows us to evaluate both performance under limited clean trajectories and whether the pre-trained model can continuously benefit from more downstream robot trajectories. Moreover, DOMINO evaluates dynamic manipulation with moving objects and changing environments, targeting the spatiotemporal reasoning abilities often missed by static manipulation benchmarks. Additional results are shown in Appendix~\ref{sec:add_domino}.

\mypara{RoboTwin~2.0 results.}
Table~\ref{tab:robotwin} shows that VLAct establishes the strongest results among the compared methods in the \emph{Base} setting and remains competitive with recent frontier VLA and world-action-model systems in the \emph{Data Scaling} setting. In the Base setting, VLAct-OFT reaches 80.5\% success with only 50 clean trajectories per task, substantially improving over the Qwen3VL-OFT baseline (61.7\%) and outperforming the compared published methods. This demonstrates that VLAct enables strong sample-efficient downstream adaptation from limited task-specific data.
In the \emph{Data Scaling} setting, VLAct-OFT reaches 92.5\% on Clean and 90.8\% on Random, outperforming large-scale systems including InternVLA-A1, Being-H0.7, Motus, LingBot-VLA, ABot-M0, and $\pi_{0.5}$. It also remains close to HoloBrain-0 and Fast-WAM. These comparisons position VLAct among the strongest reported RoboTwin~2.0 policies while avoiding a claim of absolute state of the art.
VLAct remains strong as more downstream data is added, showing that the learned representation is not merely specialized to a low-data regime. Across downstream head choices, performance remains stable, supporting the head-transfer claim in Sec.~\ref{sec:method-align}: the transferred object is the backbone representation, rather than a particular action head. Indeed, among the VLAct variants, the PI head attains the highest Clean success (93.0\%), slightly above the OFT head (92.5\%) that we use as the default configuration elsewhere in the paper; all three heads remain within 3.4 points of each other.

\begin{table}[t]
    \centering
    \small
    \setlength{\tabcolsep}{3.2pt}
    \caption{\textbf{RoboTwin~2.0 results.} Task success rate (\%) under the Base and Data Scaling settings. Base uses 50 clean trajectories per task and is evaluated on the Clean regime; Data Scaling additionally uses 500 randomized trajectories per task and is evaluated on both the Clean and Random regimes. Each reported regime evaluates 50 tasks with 100 episodes per task. Published numbers are taken from the cited papers or their same-protocol re-evaluations. Dark blue highlights our default VLAct-OFT configuration and light blue highlights alternative VLAct heads. Best results are bold and second-best results are underlined.}
    \label{tab:robotwin}
    \begin{tabular}{llccc}
    \toprule
    \multirow{2}{*}{\textbf{Method}}
    & \multirow{2}{*}{\textbf{Family}}
    & \multicolumn{1}{c}{\textbf{Base Setting}}
    & \multicolumn{2}{c}{\textbf{Data Scaling Setting}} \\
    \cmidrule(lr){3-3} \cmidrule(lr){4-5}
    & & \textbf{Clean} & \textbf{Clean} & \textbf{Random} \\
    
    \midrule
    \multicolumn{5}{l}{\emph{Published VLA and world-action-model systems}} \\
    $\pi_0$~\citep{black2024pi0}
    & Flow
    & 46.4 & 65.9 & 58.4 \\
    $\pi_{0.5}$~\citep{intelligence2025pi_,wu2026pragmatic}
    & Flow
    & 60.2 & 82.7 & 76.8 \\
    X-VLA~\citep{zheng2025x,bi2025motus}
    & Flow
    & 70.0 & 72.8 & 72.8 \\
    Lingbot-VLA~\citep{wu2026pragmatic}
    & Flow
    & -- & 88.6 & 86.7 \\
    Abot-M0~\citep{yang2026abot}
    & AML
    & -- & 86.1 & 85.1 \\
    InternVLA-A1~\citep{cai2026internvlaa1}
    & Flow
    & -- & 89.4 & 89.6 \\
    Being-H0.7~\citep{luo2026beingh07}
    & Flow
    & -- & 90.2 & 89.6 \\
    Motus~\citep{bi2025motus}
    & WAM
    & -- & 88.7 & 87.0 \\
    Fast-WAM~\citep{yuan2026fast}
    & WAM
    & -- & 91.9 & \underline{91.8} \\
    HoloBrain-0-QW~\citep{lin2026holobrain}
    & Diff.
    & -- & 91.9 & \textbf{92.3} \\

    \midrule
    \multicolumn{5}{l}{\emph{In-house baseline}} \\
    Qwen3VL-OFT 
    & OFT 
    & 61.7 & 88.2 & 88.3 \\
    
    \midrule
    \multicolumn{5}{l}{\emph{VLAct (same backbone, varying continuous head)}} \\
    \rowcolor{tableBest}
    \textbf{VLAct} 
    & OFT 
    & \textbf{80.5} & \underline{92.5} & 90.8 \\
    \rowcolor{tableSecond}
    \textbf{VLAct}
    & GR00T
    & 76.0 & 89.6 & 87.4 \\
    \rowcolor{tableSecond}
    \textbf{VLAct}
    & PI
    & \underline{77.0} & \textbf{93.0} & 88.8 \\
    
    \bottomrule
    \end{tabular}
     \vspace{-0.3cm}
\end{table}

\subsection{Real-world Experiments}
\label{sec:eval_real}


In what follows, we perform the real-world robot experiments using Franka Research 3 7-DoF arms to validate the effectiveness of our VLAct. We evaluate the proposed VLAct and a competitive baseline (Qwen3VL-4B-OFT without pre-training) across four regimes: 
\textbf{(1)} \textit{In-Domain (ID) single-arm short-horizon tasks}, such as `Carrot-from-pot', `Button pressing', `Cube stacking', `Pen in cup'; 
\textbf{(2)} \textit{ID single-arm long-horizon tasks}, \emph{e.g.}, `Table cleaning' \& `Scoop beans'; 
\textbf{(3)} \textit{ID dual-arm coordination}, with tasks `Unplugging', `Breakfast preparation', `Banana handover-place', `Fold pants', `Fold towel'; 
and 
\textbf{(4)} \textit{Out-of-Domain (OOD) generalization}, including novel objects from pot, novel objects in cup, and table cleaning with extended or fully substituted object types. 

We train two separate models for their respective evaluations: one for all single-arm tasks while another for all dual-arm tasks. Each model is trained with 50k training steps on 8 H800 GPUs, and evaluated over 10 trials to calculate the success rate. Detailed hardware settings, prompts, visualizations of each task, and success rate calculation criterion are given in Appendix~\ref{app:Real-world-exp}.

\mypara{Results.}
As shown in Fig.~\ref{fig:robocasa-gr1-vlact}, VLAct consistently improves real-world performance over the non-pre-trained Qwen3VL-4B-OFT baseline.
In single-arm short-horizon ID tasks (Fig.~\ref{fig:robocasa-gr1-vlact} (a)), VLAct achieves 92.5\% average success, higher than the baseline's 77.5\%, with clear gains on spatially sensitive tasks such as cube stacking and pen-in-cup. VLAct also generalizes better to novel objects: on the short-horizon OOD tasks, it obtains 90.0\% success on both `novel object from pot' and `novel object in cup', compared with 73.3\% and 65.0\% for the baseline. In single-arm long-horizon tasks (Fig.~\ref{fig:robocasa-gr1-vlact} (b)), VLAct achieves weighted scores of 86.6\% on table cleaning and 80.0\% on scooping beans, outperforming the baseline scores of 73.3\% and 33.3\%. Under harder OOD long-horizon settings, VLAct keeps high scores of 82.5\% for extended-sequence table cleaning and 83.3\% for full object substitution, while the baseline drops to 47.5\% and 46.6\%. Finally, despite being pre-trained only on single-arm data, VLAct transfers effectively to dual-arm coordination, reaching 72.0\% average success across dual-arm tasks versus 44.0\% for the baseline, with the largest gains on deformable-object and synchronization-heavy tasks such as pants folding and towel folding (Fig.~\ref{fig:robocasa-gr1-vlact} (c)). 
These results suggest that VLAct improves not only in-domain manipulation accuracy, but also novel-object generalization, long-horizon execution, and dual-arm coordination.

%% file: RAW/sections/5_analysis.tex
\subsection{Does VLAct Transfer to Unseen Embodiments?}
\label{sec:exp-cross}

\mypara{Setup.}
We test whether VLAct learns action representations that transfer to embodiments unseen during continued pre-training. VLAct is continually pre-trained only on Franka single-arm and AgileX bimanual data; both the GR-1 humanoid and the ARX X5 bimanual platform are held out during continued pre-training. We fine-tune VLAct-OFT on RoboCasa-GR1 under a fixed downstream protocol and additionally evaluate VLAct through the official RoboDojo simulation benchmark on ARX X5. These settings directly probe cross-embodiment transfer because the target robots differ from the pre-training embodiments in morphology, action space, and manipulation dynamics.

\begin{figure}[t]
    \centering
    \vspace{-0.5cm}
    \includegraphics[width=1.0\linewidth]{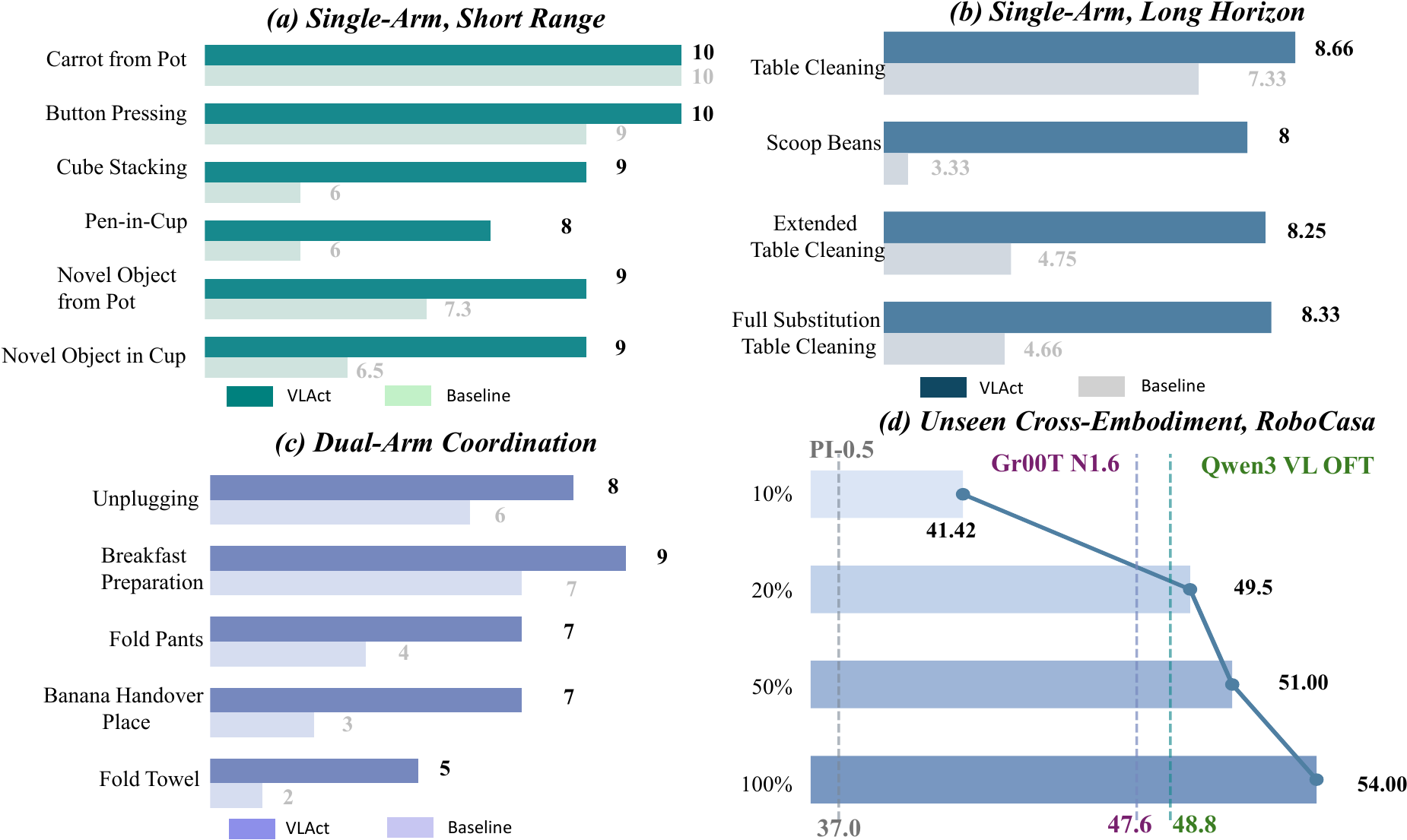}
    \vspace{-0.5cm}
    \caption{\textbf{Real-machine evaluation and cross-embodiment transfer.}
    (a)--(c) Real-machine evaluation results on single-arm short-range, single-arm long-horizon, and dual-arm coordination tasks, where VLAct consistently improves over the baseline across diverse manipulation settings. (d) Cross-embodiment transfer on RoboCasa-GR1, where VLAct transfers to the held-out GR-1 embodiment and surpasses reported full-data baselines with only $20\%$ downstream data.}

    \label{fig:robocasa-gr1-vlact}
\end{figure}

\mypara{RoboCasa-GR1 results.}
Fig.~\ref{fig:robocasa-gr1-vlact}(d) reports RoboCasa-GR1 performance under different fractions of the downstream training data. With full fine-tuning data, VLAct reaches \textbf{54.0\%}, outperforming the strongest reported full-data baselines, including Qwen3VL-OFT at $48.8\%$, GR00T-N1.6 at $47.6\%$, and $\pi_{0.5}$ at $37.0\%$. Since GR-1 trajectories are never used in VLAct pre-training, this result shows that the backbone learned from Franka and AgileX can provide a stronger initialization for a new humanoid embodiment.

The data-efficiency curve further supports this conclusion. With only $20\%$ of the RoboCasa-GR1 trajectories, VLAct already reaches $49.5\%$, matching or exceeding the full-data Qwen3VL-OFT baseline. With $50\%$ of the data, performance rises to $51.0\%$, and full-data fine-tuning further improves it to $54.0\%$. Thus, VLAct is not merely benefiting from downstream data scale; the pretrained backbone transfers useful action structure to the unseen embodiment before full task-specific adaptation. These results support the central claim that VLAct learns reusable action representations rather than overfitting to the robot embodiments observed during pre-training.

\mypara{RoboDojo results.}
RoboDojo~\citep{chen2026robodojo} evaluates policies on 42 ARX X5 simulation tasks spanning Generalization, Precision, Long-Horizon, Memory, and Open capabilities. Each task is evaluated over 50 episodes, and the benchmark reports both a partial-progress score and binary task success. Table~\ref{tab:robodojo} reports the \href{https://robodojo-benchmark.com/leaderboard}{official leaderboard} snapshot from August~24, 2026. Among 35 policies, VLAct ranks \textbf{eighth by average score} and \textbf{sixth by average success rate}, placing it in the top quartile under both metrics. Among the four explicitly designated world-action-model (WAM) entries, X-WAM is the strongest, with a 7.69 score and 3.83\% success rate. VLAct exceeds X-WAM by \textbf{2.97 score points} and \textbf{3.77 percentage points} in success, thereby outperforming every designated WAM entry on both aggregate metrics. VLAct also outperforms several industry-developed systems, including Xiaomi-Robotics-0, GalaxeaVLA~(G0), LingBot-VLA~\citep{wu2026pragmatic}, and ABot-M0~\citep{yang2026abot}. Moreover, six of the seven higher-scoring entries are submitted by industry teams or robotics companies: Dexmal, Galaxea AI, Xiaomi Robotics, Tencent Robotics X, OpenHelix Robotics, and Physical Intelligence. Although the leaderboard does not normalize for training compute, this result is notable given that VLAct uses fully open-source data and a 16-GPU continued pre-training setup.

Compared with the Qwen3-VL-based StarVLA-$\alpha$~\citep{ye2026starvla} entry, VLAct improves the average score by 4.26 points and success by 4.36 percentage points. The largest gains occur on Precision and Long-Horizon tasks, while Memory remains a clear limitation. Since ARX X5 is also absent from VLAct pre-training, these results provide an additional cross-embodiment test under a broad, externally defined evaluation suite.

\begin{table*}[t]
    \centering
    \scriptsize
    \setlength{\tabcolsep}{3pt}
    \renewcommand{\arraystretch}{0.94}
    \caption{\textbf{RoboDojo simulation results.} Top 20 policies from the official August~24, 2026 leaderboard, ordered by average score. Each cell reports partial-progress score / success rate (\%). Gen.-Std. and Gen.-Rand. denote the standard and randomized Generalization settings. The full leaderboard contains 35 policies; VLAct ranks eighth by score and sixth by success rate. The \colorbox{tableBest}{dark-blue row} highlights VLAct; \colorbox{tableSecond}{light-blue rows} and \textsuperscript{$\dagger$} identify explicitly designated WAM entries.}
    \label{tab:robodojo}
    \resizebox{\textwidth}{!}{%
    \begin{tabular}{lccccccc}
    \toprule
    \textbf{Method} & \textbf{Gen.-Std.} & \textbf{Gen.-Rand.} & \textbf{Precision} & \textbf{Long-Horizon} & \textbf{Memory} & \textbf{Open} & \textbf{Average} \\
    \midrule
    DM0.5~\citep{dexmal2026dm05} & 23.49 / 18.00 & 8.06 / 4.00 & 24.82 / 16.75 & 33.70 / 19.50 & 47.74 / 47.44 & 2.43 / 2.08 & 24.90 / 19.34 \\
    GalaxeaVLA~(G0.5)~\citep{liu2026g05} & 26.74 / 20.00 & 11.16 / 6.00 & 28.25 / 20.42 & 44.12 / 32.25 & 8.61 / 7.33 & 1.73 / 1.58 & 20.23 / 14.88 \\
    Xiaomi-Robotics-1~\citep{team2026xiaomi} & 35.65 / 28.00 & 11.44 / 6.00 & 26.69 / 18.83 & 38.39 / 23.67 & 7.81 / 6.56 & 3.94 / 3.58 & 20.07 / 13.93 \\
    Hy-Embodied-0.5-VLA~\citep{zhang2026hy} & 21.98 / 17.00 & 1.57 / 0.00 & 13.81 / 8.00 & 25.74 / 14.92 & 13.37 / 12.11 & 0.65 / 0.58 & 13.07 / 8.80 \\
    Spatial Forcing~\citep{li2025spatial} & 21.25 / 15.00 & 6.98 / 4.00 & 17.33 / 10.58 & 23.26 / 14.58 & 5.43 / 4.11 & 1.78 / 1.58 & 12.38 / 8.04 \\
    $\pi_{0.5}$~\citep{intelligence2025pi_} & 20.93 / 15.00 & 5.82 / 1.00 & 12.40 / 5.50 & 23.54 / 14.67 & 5.78 / 4.56 & 1.98 / 1.67 & 11.41 / 6.91 \\
    InternVLA-A1.5~\citep{ma2026internvlaa15} & 16.81 / 12.00 & 3.90 / 2.00 & 15.23 / 10.17 & 23.80 / 13.75 & 4.93 / 3.56 & 1.43 / 1.42 & 11.15 / 7.14 \\
    \rowcolor{tableBest}
    \textbf{VLAct} & 16.33 / 12.00 & 2.74 / 1.00 & 20.62 / 15.25 & 20.12 / 13.67 & 0.66 / 0.56 & 2.37 / 2.25 & \textbf{10.66 / 7.60} \\
    X-VLA~\citep{zheng2025x} & 17.90 / 12.00 & 3.04 / 1.00 & 18.32 / 12.00 & 16.53 / 9.75 & 4.76 / 3.56 & 0.55 / 0.50 & 10.13 / 6.52 \\
    \rowcolor{tableSecond}
    X-WAM\textsuperscript{$\dagger$}~\citep{guo2026unified4dworldaction} & 11.24 / 5.00 & 3.54 / 1.00 & 6.72 / 1.83 & 17.47 / 9.08 & 6.32 / 4.67 & 0.57 / 0.25 & 7.69 / 3.83 \\
    Xiaomi-Robotics-0~\citep{cai2026xiaomi} & 13.81 / 11.00 & 1.05 / 0.00 & 8.42 / 4.58 & 13.51 / 6.92 & 5.07 / 3.67 & 0.22 / 0.17 & 6.93 / 4.18 \\
    StarVLA-$\alpha$~\citep{ye2026starvla} & 7.54 / 5.00 & 0.33 / 0.00 & 9.90 / 4.33 & 14.15 / 6.50 & 3.34 / 2.44 & 0.68 / 0.58 & 6.40 / 3.24 \\
    \rowcolor{tableSecond}
    GigaWorld-Policy-0\textsuperscript{$\dagger$}~\citep{ye2026gigaworld} & 10.28 / 6.00 & 0.41 / 0.00 & 6.15 / 1.83 & 15.51 / 8.92 & 3.46 / 2.22 & 0.54 / 0.50 & 6.20 / 3.27 \\
    GalaxeaVLA~(G0)~\citep{jiang2025galaxea} & 8.71 / 6.00 & 0.36 / 0.00 & 8.10 / 3.83 & 12.60 / 5.58 & 3.17 / 1.89 & 0.70 / 0.67 & 5.82 / 2.96 \\
    LingBot-VLA~\citep{wu2026pragmatic} & 10.88 / 8.00 & 2.55 / 1.00 & 5.33 / 1.83 & 10.89 / 5.25 & 3.82 / 2.78 & 0.72 / 0.67 & 5.50 / 2.96 \\
    EventVLA~\citep{yang2026eventvla} & 6.68 / 3.00 & 1.22 / 0.00 & 10.13 / 5.75 & 5.05 / 0.83 & 4.92 / 4.78 & 0.80 / 0.75 & 4.97 / 2.81 \\
    \rowcolor{tableSecond}
    AHA-WAM\textsuperscript{$\dagger$}~\citep{cai2026aha} & 10.32 / 6.00 & 1.26 / 0.00 & 5.86 / 2.42 & 8.61 / 2.67 & 2.97 / 2.78 & 0.88 / 0.83 & 4.82 / 2.39 \\
    ABot-M0~\citep{yang2026abot} & 9.20 / 5.00 & 2.26 / 2.00 & 5.50 / 1.75 & 3.96 / 0.50 & 2.44 / 2.22 & 0.72 / 0.67 & 3.67 / 1.73 \\
    \rowcolor{tableSecond}
    Fast-WAM\textsuperscript{$\dagger$}~\citep{yuan2026fast} & 4.33 / 2.00 & 0.34 / 0.00 & 1.96 / 0.00 & 9.14 / 5.17 & 3.55 / 3.44 & 0.42 / 0.42 & 3.48 / 2.03 \\
    $\pi_0$~\citep{black2024pi0} & 7.18 / 5.00 & 0.71 / 0.00 & 3.56 / 0.75 & 6.19 / 2.00 & 3.47 / 2.11 & 0.25 / 0.25 & 3.48 / 1.53 \\
    \bottomrule
    \end{tabular}%
    }

\end{table*}

%

%% file: RAW/sections/6_related_work.tex
\section{Related Work}
\label{sec:related}

\mypara{Generalist Robot Policies.}
Recent research has shifted toward generalist policies via Vision-Language-Action models~\cite{brohan2022rt, kim2024openvla, zitkovich2023rt}. By fine-tuning web-scale vision-language models~\cite{beyer2024paligemma} on diverse robotic datasets~\cite{o2024open, khazatsky2024droid}, VLAs exhibit remarkable generalization across novel instructions and environments. However, frontier models (e.g., the $\pi$ series~\cite{black2024pi0}) often rely on closed-source data, hindering reproducibility. Our work provide a systematic recipe study using only public data, focusing on the representation learning nuances that determine VLA performance. The additional related work is shown in Appendix~\ref{app:related_work}.

%% file: RAW/sections/7_conclusion.tex
\section{Conclusion}

We presented \textbf{VLAct}, a representation-centric VLA continued pre-training recipe for building more transferable VLM backbones from limited robot data. Instead of treating robot continued pre-training as large-scale action fitting, VLAct is designed to preserve the broad VLM prior, avoid over-specialization to a single action head, and encourage shared action semantics across embodiments.
Across multi-embodiment simulation benchmarks, real-world robot experiments, and unseen-embodiment transfer, VLAct consistently improves downstream performance under fixed fine-tuning protocols. It surpasses strong VLA baselines on LIBERO-Plus, achieves state-of-the-art performance on RoboTwin~2.0, ranks in the top quartile of the all policy RoboDojo leaderboard while outperforming several industry-developed systems, and rapidly adapts to a humanoid embodiment unseen during continued pre-training on RoboCasa-GR1.
These results are obtained using fully open-source data and only a 16-GPU training setup. Their competitiveness against predominantly industry-backed systems shows that VLA progress is not only a matter of scaling robot trajectories or compute: better continued pre-training design can substantially improve what a fixed resource budget delivers.

%% file: RAW/sections/Authors.tex
\section*{Authors}
\addcontentsline{toc}{section}{Authors}

\begin{tcolorbox}[
  enhanced,
  colback=tableSecond!28!white,
  colframe=metablue!85!black,
  arc=4pt,
  boxrule=0.6pt,
  left=12pt, right=12pt,
  top=10pt, bottom=10pt,
  drop shadow=gray!20!white
]
\begin{minipage}[t]{0.48\linewidth}
\vspace{0pt}
\raggedright
{\large\sffamily\bfseries Core Contributors}
\begin{itemize}[noitemsep,topsep=5pt,leftmargin=1.4em]
  \item Senqiao Yang\textsuperscript{\textcolor{metablue}{\ensuremath{\dagger}}}
  \item Chengyao Wang\textsuperscript{\textcolor{metablue}{\ensuremath{\dagger}}}
  \item Yuxin Chen
  \item Zixuan Wang
  \item Longxiang Tang
  \item Haokun Gui
  \item Jinhui Ye
  \item Changsheng Lu
  \item Xiaoyang Wu
  \item Mingkang Zhu
\end{itemize}
\end{minipage}\hfill
\begin{minipage}[t]{0.44\linewidth}
\vspace{0pt}
\raggedright
{\large\sffamily\bfseries Advisors}
\begin{itemize}[noitemsep,topsep=5pt,leftmargin=1.4em]
  \item Pengguang Chen
  \item Shu Liu\textsuperscript{\textcolor{metablue}{\ding{41}}}
  \item Zhuotao Tian
  \item Hengshuang Zhao
  \item Bei Yu
  \item Jiaya Jia
\end{itemize}
\end{minipage}
\end{tcolorbox}

\medskip
\begin{center}
{\small
\textsuperscript{\textcolor{metablue}{\ensuremath{\dagger}}} Project leaders
\qquad
\textsuperscript{\textcolor{metablue}{\ding{41}}} Correspondence
}\par\vspace{6pt}
{\small For questions or suggestions, please feel free to contact us at \texttt{yangsenqiao.ai@gmail.com}.}
\end{center}

%% file: RAW/sections/X_appendix.tex
\appendix

\section{Additional Related Work}
\label{app:related_work}
\mypara{Generalist Robot Policies and Data Pretraining}

The emergence of foundation models~\cite{bommasani2021opportunities} has catalyzed a shift from task-specific controllers to generalist robotic policies. Early approaches often deploy pre-trained models as high-level semantic planners that interface with low-level, robot-specific skills~\cite{zitkovich2023rt, driess2023palm, huang2022inner, liang2023code, lin2023text2motion, singh2022progprompt}. To overcome the hierarchy's rigidity, Vision-Language-Action (VLA) models have gained prominence by fine-tuning backbones such as Eagle-2~\cite{li2025eagle} or PaliGemma~\cite{beyer2024paligemma}directly on embodied data for end-to-end optimization ~\cite{black2024pi0, zitkovich2023rt, bjorck2025gr00t, kim2024openvla, yang2026abot,team2025gemini,ye2026starvla,community2026starvla}. Modern VLA architectures frequently incorporate advanced generation techniques like flow-matching~\cite{lipman2022flow, liu2022flow} and action chunking~\cite{zhao2023learning}. While some frameworks use Mixture-of-Experts to bridge vision and action~\cite{black2024pi0}, others utilize cross-attention or embodiment-specific projectors to facilitate multi-robot transfer across latent~\cite{team2024octo}.

The performance of these models is fundamentally driven by data scaling. While high-fidelity trajectories from teleoperation~\cite{aldaco2024aloha, mandlekar2018roboturk, mandlekar2019scaling, wu2024gello} have been scaled into massive datasets with thousands of hours of experience~\cite{bu2025agibot, black2024pi0, brohan2022rt, walke2023bridgedata, lynch2023interactive, o2024open}, the high cost of collection remains a bottleneck. Consequently, alternative sources such as instrumented human trajectories~\cite{chi2024universal, fang2024airexo} and internet-scale human video datasets~\cite{damen2020epic, grauman2022ego4d} are increasingly leveraged. These non-robotic sources serve either as a substrate for representation pretraining~\cite{karamcheti2023language, wu2023unleashing} or as a means of joint learning through motion-based intermediate representations~\cite{bharadhwaj2024roboagent, ye2024latent}. In contrast to prior efforts that often rely on proprietary training stacks, our work systematically explores the nuances of learning from a diverse mixture of public robot data, human videos, and synthetic samples.

\mypara{Cross-task and cross-embodiment generalization}
Achieving robust generalization across diverse environments, tasks, and embodiments is a central objective in robot learning. A prevalent strategy involves leveraging human video data to bootstrap policies, either via general representation learning~\cite{nair2022r3m,bhateja2023robotic}, direct supervision through human motion cloning~\cite{shaw2023videodex,bharadhwaj2023zero,kareer2025egomimic} or the extraction of 2D point tracks as intermediate representations~\cite{bharadhwaj2024track2act,vecerik2024robotap,wen2023any}. Parallel to these data-driven efforts, another line of work integrates Internet-scale foundation models into training or inference pipelines to inherit broad semantic knowledge~\cite{kapelyukh2023dall,yu2023scaling,jiang2023vima}. With the emergence of large-scale, cross-embodiment datasets~\cite{o2024open}, recent research has increasingly focused on enhancing transfer learning between heterogeneous robot platforms~\cite{doshi2024scaling,yang2024pushing}. 

\mypara{Vision Language Models}
Modern Vision-Language Models (VLMs) ~\cite{chen2024lion,liu2024improved,dai2023instructblip,bai2025qwen3,li2024monkey,wang2024qwen2,yang2025visionzip,yang2026mage} have transitioned from task-specific designs toward unified frameworks capable of general-purpose multimodal reasoning~\cite{li2024optimus,ye2024cat,yang2023lisa++}. Typically comprising a vision encoder, a projector, and a Large Language Model (LLM) backbone~\cite{liu2024improved,wang2024qwen2}, these models unify diverse tasks as textual outputs. This shift is driven by visual instruction tuning~\cite{liu2023visual,chen2024expanding}, which grants VLMs the flexibility to follow open-ended commands and generalize to out-of-distribution scenarios. Recent advancements further push these boundaries through dynamic resolution~\cite{bai2025qwen3}, unified image-video processing~\cite{li2024llava}, and reinforcement-learning-enhanced reasoning~\cite{huang2025vision,yang2026visionthink}, significantly bolstering real-world perception. While VLMs have demonstrated high-level planning in domains like autonomous driving~\cite{tian2024drivevlm} and GUI interaction~\cite{hong2024cogagent}, they often remain passive observers. The emerging frontier lies in bridging this "embodiment gap" by extending VLM reasoning into direct physical interaction. By grounding these foundation models into action spaces, Vision-Language-Action (VLA) models leverage the rich semantic and spatial knowledge of VLMs to enable robust robotic manipulation in complex environments.

\section{Additional Main Results}
\subsection{Additional Results on Franka}
\label{sec:add_franka}

\mypara{Results on VLA-Arena.}
Table~\ref{tab:app-vla-arena} reports success on VLA-Arena across four behavioral generalization axes. VLAct achieves the best score on every axis and reaches \textbf{54.8\%} overall. 
Compared to Qwen3-VL-OFT, our VLAct improves the average score by \textbf{21.4\%} despite using the same backbone family and downstream OFT head. This large gap again isolates the effect of the representation-centric pre-training recipe rather than the base VLM or the fine-tuning head.

VLAct also outperforms the strongest public baseline, $\pi_{0.5}$, by 10.5 points overall, with especially large gains on \emph{Long-Horizon} and \emph{Safety}, 21.0 and 11.7 points, respectively. Since all methods are fine-tuned on the same downstream task data, the improvement directly demonstrates the effectiveness of VLAct as a pretrained VLA backbone: it provides a substantially stronger starting point for downstream policy learning and yields better generalization across challenging behavioral settings.

\begin{table}[t]
\centering
\small
\setlength{\tabcolsep}{5pt}
\caption{\textbf{VLA-Arena results.} Success rate (\%) averaged over difficulty levels $\mathrm{L}0$/$\mathrm{L}1$/$\mathrm{L}2$ within each task category. The final average follows the official suite weighting over $11$ task suites. Dark blue highlights VLAct and per-column best results; light blue highlights the strongest public baseline by average and per-column second-best results.}
\label{tab:app-vla-arena}
\begin{tabular}{lccccc}
\toprule
\textbf{Method} & \textbf{Safety} & \textbf{Distractor} & \textbf{Extrap.} & \textbf{Long-H.} & \textbf{Avg.} \\
\midrule
SmolVLA~\citep{shukor2025smolvla}             & 16.5 & 21.0 & 10.0 & 24.7 & 16.3 \\
$\pi_0$-FAST~\citep{pertsch2025fast}          & 34.1 & 39.0 &  4.2 & 20.7 & 25.6 \\
GR00T-N1.6~\citep{bjorck2025gr00t}            & 32.8 & 40.7 & 16.7 & 10.3 & 27.8 \\
Qwen3-VL-$\pi$~\citep{bai2025qwen3}      & 38.3 & 45.3 & 22.4 & 25.3 & 34.1 \\
UniVLA~\citep{wang2025unified}                   & 45.5 & 41.3 & 31.1 & 22.0 & 38.7 \\
OpenVLA~\citep{kim2024openvla}                & 41.9 & 43.0 & \secondcell{36.7} & 26.7 & 39.3 \\
OpenVLA-OFT~\citep{kim2025fine}                & 43.2 & 52.3 & 30.4 & 26.7 & 39.9 \\
$\pi_0$~\citep{black2024pi0}                  & 49.7 & 43.7 & 32.7 & 31.3 & 42.3 \\
\rowcolor{tableSecond}
$\pi_{0.5}$~\citep{intelligence2025pi_}      & \secondcell{51.5} & \secondcell{54.0} & 31.1 & 29.0 & \secondcell{44.3} \\
Qwen3-VL-OFT                                  & 36.1 & 40.0 & 24.7 & \secondcell{32.7} & 33.4 \\
\midrule
\rowcolor{tableBest}
\textbf{VLAct}                                & \bestcell{63.2} & \bestcell{64.0} & \bestcell{36.9} & \bestcell{50.0} & \bestcell{54.8} \\
\bottomrule
\end{tabular}
\end{table}

\subsection{Additional Results on Agilex}
\label{sec:add_domino}
\mypara{Results on DOMINO.}
Table~\ref{tab:app-domino} reports results on DOMINO, a dynamic manipulation benchmark with moving objects and changing scene states. VLAct-OFT achieves the best performance on both metrics, reaching \textbf{18.50} SR and \textbf{34.20} MS. Compared with Qwen3VL-OFT, the strongest baseline with the same Qwen3-VL backbone family and OFT downstream head, VLAct improves SR by \textbf{7.64} points and MS by 3.71 points. Since both models are fine-tuned under the same downstream setting, the improvement highlights the benefit of VLAct as a pretrained backbone for dynamic bimanual manipulation. Together, the RoboTwin~2.0 and DOMINO results show that VLAct is effective across both visually randomized static manipulation and temporally evolving dynamic manipulation.
\begin{table}[t]
\centering
\small
\setlength{\tabcolsep}{6pt}
\caption{\textbf{DOMINO dynamic manipulation results.}
We evaluate one policy on all 35 DOMINO dynamic manipulation tasks under the clean dynamic setting. DOMINO reports Success Rate (SR) and Manipulation Score (MS). Dark blue highlights VLAct-OFT; light blue highlights the strongest baseline.}
\label{tab:app-domino}
\begin{tabular}{llcc}
\toprule
\textbf{Model} & \textbf{Backbone} & \textbf{SR $\uparrow$} & \textbf{MS $\uparrow$} \\
\midrule
OpenVLA & Llama-2 & 1.54 & 6.10 \\
RDT-1B & DiT-1B & 5.34 & 17.71 \\
$\pi_0$ & PaliGemma & 8.17 & 23.96 \\
$\pi_0$-FAST & PaliGemma & 3.54 & 20.87 \\
$\pi_{0.5}$ & PaliGemma & 9.63 & 26.17 \\
InternVLA-M1 & InternVL & 5.40 & 27.57 \\
OpenVLA-OFT & Llama-2 & 9.06 & 24.06 \\
\rowcolor{tableSecond}
Qwen3VL-OFT & Qwen3-VL & \secondcell{10.86} & \secondcell{30.49} \\
\midrule
\rowcolor{tableBest}
\textbf{VLAct-OFT} & Qwen3-VL & \bestcell{18.50} & \bestcell{34.20} \\
\bottomrule
\end{tabular}
\end{table}

\section{Effectiveness of Shallow-Layer Protection During Pre-training}
\label{sec:freeze_detail}

\subsection{Motivation and Design}
\label{sec:freeze_motivation}

Our VLA-oriented backbone is initialized from a VLM trained on broad vision-language data. This initialization already contains useful low-level visual features, spatial cues, and early vision-language alignment. In contrast, VLA pre-training data is much narrower: robot trajectories often come from fixed camera viewpoints, repeated manipulation scenes, and embodiment-specific action correlations. If all layers are updated end-to-end during VLA pre-training, gradients from this narrow distribution can shift the backbone away from its general VLM representation before the model has learned robust action-conditioned features.

\begin{figure}[t]
    \centering
    \includegraphics[width=\linewidth]{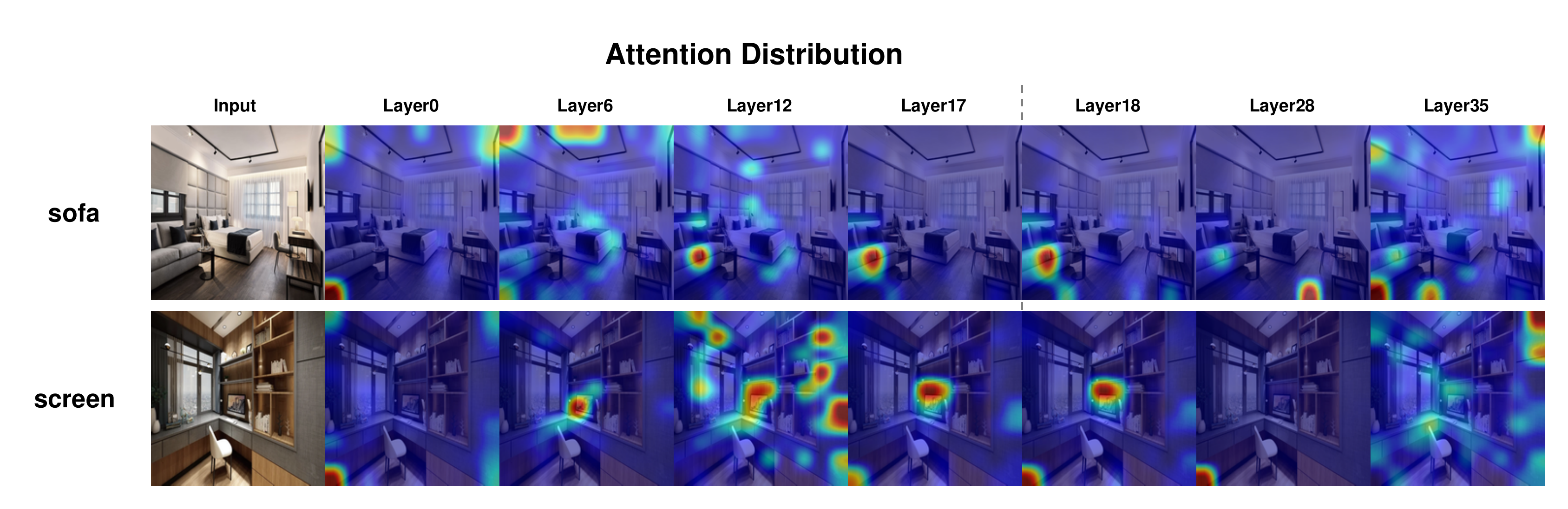}
    \caption{\textbf{Layer-wise attention visualization.}
    Lower layers attend more broadly to visual and spatial information, while deeper layers focus more on semantic and task-relevant regions. This motivates freezing the vision encoder and lower LLM layers during VLA pre-training to reduce representation drift under limited robot data.}
    \label{fig:shallow_layer_attention}
\end{figure}

Shallow-layer protection is designed to reduce this representation drift. As illustrated in Fig.~\ref{fig:shallow_layer_attention}, lower layers tend to preserve broad visual and spatial information, while deeper layers focus more on semantic and task-relevant regions. We therefore freeze the entire vision encoder and the lower half of the LLM layers during VLA pre-training. This protects the general perceptual and early alignment features inherited from the VLM, while leaving the upper LLM layers and action heads trainable for instruction following, task semantics, and action-conditioned reasoning. During downstream fine-tuning, we unfreeze the full model so that all parameters can adapt to the target task and embodiment.

\subsection{Ablation Results}
\label{sec:freeze_ablation_results}

Table~\ref{tab:freeze_ablation} reports the ablation. Updating the full backbone during VLA pre-training gives weaker downstream performance, with 78.9\% on LIBERO-Plus and 77.1\% on RoboTwin2.0. Freezing only the vision encoder improves performance to 81.3 and 79.3\%, showing that protecting visual features is already helpful. Freezing both the vision encoder and the lower half of the LLM performs best, reaching 82.6\% on LIBERO-Plus and 80.5\% on RoboTwin2.0, improving over full-backbone updating by 3.7\% and 3.4\%, respectively. These results support the role of shallow-layer protection in our recipe: it preserves the general VLM representation during pre-training, while leaving enough upper-layer capacity for action-aware adaptation.

\begin{table}[t]
\centering
\small
\caption{\textbf{Ablation on shallow-layer protection.}
Freezing the vision encoder and lower LLM layers during VLA pre-training reduces representation drift and improves downstream fine-tuning. Dark blue highlights the full design; light blue highlights the strongest partial variant.}
\label{tab:freeze_ablation}
\begin{tabular}{lcc}
\toprule
\textbf{Pre-training update strategy} & \textbf{LIBERO-Plus} & \textbf{RoboTwin2.0} \\
\midrule
Update full backbone & 78.9 & 77.1 \\
\rowcolor{tableSecond}
Freeze vision encoder only & \secondcell{81.3} & \secondcell{79.3} \\
\rowcolor{tableBest}
Freeze vision encoder + lower 1/2 LLM & \bestcell{82.6} & \bestcell{80.5} \\
\bottomrule
\end{tabular}
\end{table}

\section{VLM Co-training for Representation Preservation}
\label{sec:vlm_data}
\subsection{Why Mix Auxiliary Data During VLA Pre-training?}
\label{sec:aux_data_motivation}

Our first design goal is to preserve the general foundation-model representation inherited from the VLM backbone while adapting it to robot control. During action-only VLA pre-training, the trainable layers are optimized through a narrow supervision channel: repeated robot scenes, embodiment-specific correlations, and sparse action targets. If used alone, this signal can make the backbone over-specialize to the robot trajectory distribution and weaken the semantic, visual, and language coverage that made the original VLM useful.

We therefore mix auxiliary non-action data into VLA pre-training. The goal is not simply to add more samples, but to provide representation-preserving supervision outside the action channel. Robot trajectories teach action-conditioned control, while auxiliary foundation-model data keeps the trainable layers closer to their original operating regime and exposes them to more diverse feature updates. Thus, auxiliary co-training acts as both a representation anchor and a source of feature diversity.

Among the auxiliary sources, image captions play the central role. Detailed captions are close to the dominant supervision used to train the original VLM and provide dense semantic gradients over objects, attributes, relations, and scene context. Grounding and spatial-QA data further constrain localized and spatial representations. We additionally include pure language instruction data as an off-domain diagnostic source: because it is not directly tied to robot perception or control, any gain from this source would suggest that the benefit of auxiliary co-training is not only task-specific transfer, but also representation preservation and diversification.

At every pre-training step, each minibatch contains both robot-trajectory samples and auxiliary samples. We optimize the action loss together with an auxiliary VLM cross-entropy loss:
$$
    \mathcal{L}_{\mathrm{total}}
    =
    \mathcal{L}_{\mathrm{action}}
    +
    0.5\,\mathcal{L}_{\mathrm{VLM\text{-}CE}} .
$$

\subsection{Auxiliary Data Mixture}
\label{sec:aux_data_sources}

We study five types of auxiliary supervision: image captioning, bounding-box question answering (BBox-QA), point question answering (Point-QA), spatial question answering (Spatial-QA), and pure language instruction data. These sources provide different constraints on the backbone representation. Caption data preserves broad semantic grounding; BBox-QA and Point-QA strengthen localized visual grounding; Spatial-QA targets spatial relations and viewpoint reasoning; pure language instruction data supplies off-domain, non-action supervision on the language side of the backbone. Table~\ref{tab:aux_data_details} summarizes the processed auxiliary datasets used in our study.

\begin{table}[h]
\centering
\small
\setlength{\tabcolsep}{8pt}
\caption{\textbf{Auxiliary data used during VLA pre-training.}
The mixture covers captioning, grounding, spatial reasoning, and pure language instruction supervision.}
\label{tab:aux_data_details}
\begin{tabular}{ll}
\toprule
\textbf{Type} & \textbf{Dataset} \\
\midrule
\multirow{2}{*}{Image Caption} 
 & LLaVA-ReCap-CC3M~\citep{li2024llavanextablations} \\
 & LLaVA OneVision~\citep{li2024llava} \\
\midrule
\multirow{2}{*}{BBox-QA} 
 & RefCOCO~\citep{chen2025revisiting} \\
 & COCO-ReM~\citep{singh2024benchmarking} \\
\midrule
\multirow{2}{*}{Point-QA} 
 & PixMo-Points~\citep{deitke2025molmo} \\
 & RoboPoint~\citep{yuan2024robopoint} \\
\midrule
Spatial-QA 
 & SenseNova-SI-800K~\citep{cai2025scaling} \\
\midrule
Pure Language Instruction
 & Nemotron-SFT-Instruction-Following-Chat-v2 \\
\bottomrule
\end{tabular}
\end{table}
\mypara{Image caption data.}
We use image caption data as the primary semantic anchor for the VLM representation. LLaVA-ReCap-CC3M~\citep{li2024llavanextablations} provides detailed recaptions of web images, preserving broad visual coverage while replacing short or noisy alt-text with richer descriptions. ShareGPT4V~\citep{chen2024sharegpt4v} provides high-quality descriptive captions that emphasize object identity, attributes, spatial relations, world knowledge, and visual details. These properties make caption data especially suitable for preserving the backbone's original vision-language representation during robot pre-training.

\mypara{Grounding data.}
BBox-QA and Point-QA data provide localized visual-language supervision. RefCOCO~\citep{chen2025revisiting} and COCO-ReM~\citep{singh2024benchmarking} are converted into bounding-box question-answering conversations, with coordinates normalized to the $0$--$1000$ convention used by the Qwen3-VL family. PixMo-Points~\citep{deitke2025molmo} and RoboPoint~\citep{yuan2024robopoint} provide point-based grounding signals. RoboPoint is particularly relevant to robotics because its supervision targets spatial affordance prediction from image-instruction pairs. For efficiency, we filter overly large images and unusually long samples, and restrict each grounding sample to at most 10 boxes or points.

\mypara{Spatial-QA data.}
SenseNova-SI-800K~\citep{cai2025scaling} provides image-conditioned conversations designed to strengthen spatial reasoning. The data covers relative positions, directions, viewpoints, and multi-image spatial context. We keep only single-image cases to control memory cost and remove unusually long conversations for training stability.

\begin{figure}[h]
\centering
\includegraphics[width=0.95\linewidth]{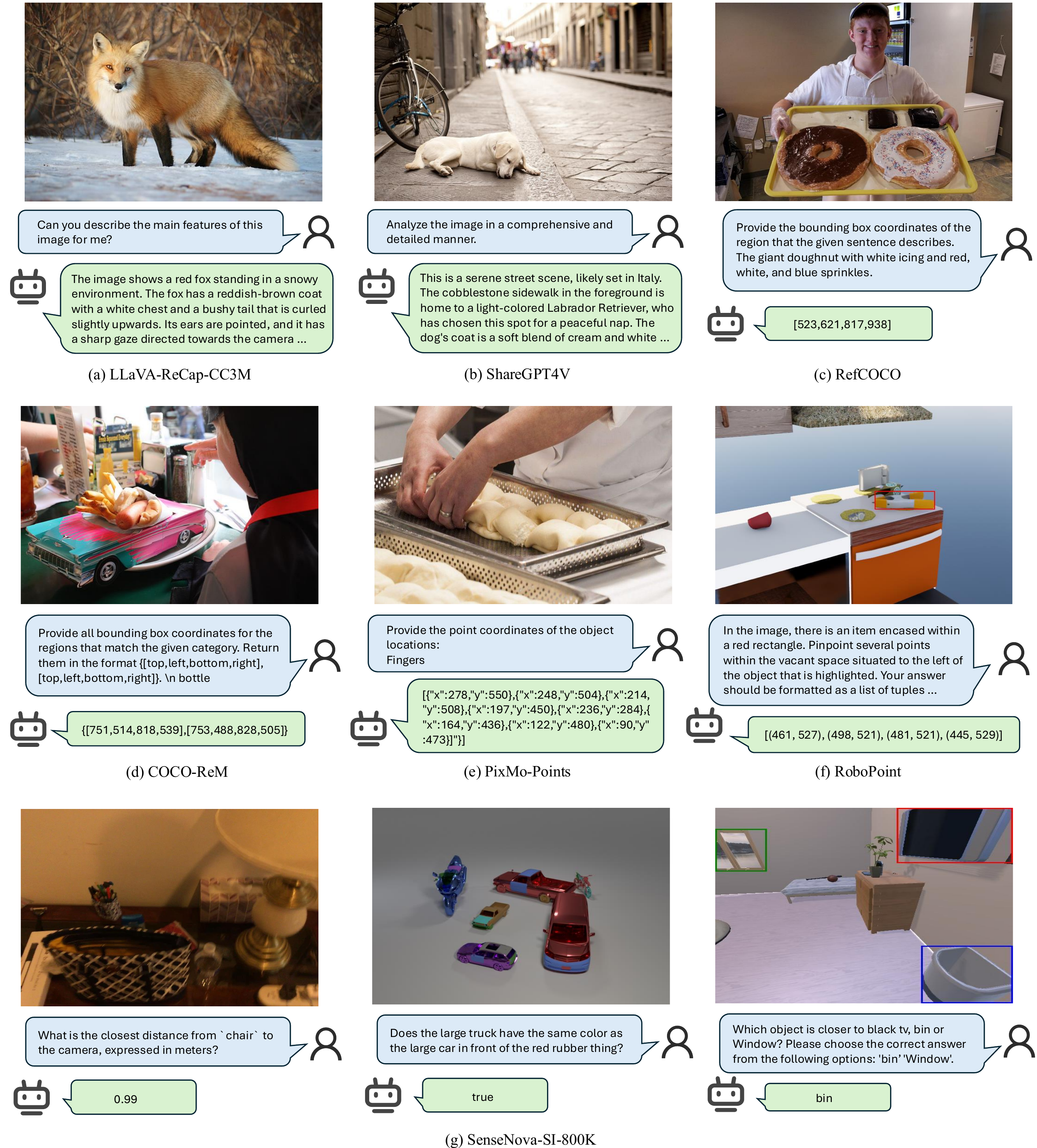}
\caption{\textbf{Examples from the processed VLM data mixture.}
We show representative samples from captioning, grounding, and spatial-reasoning datasets used during VLA pre-training.}
\label{fig:vlm_data_cases}
\end{figure}

We inspect samples from each processed data source to verify the converted format and supervision signal. Representative examples are shown in Fig.~\ref{fig:vlm_data_cases}.

\mypara{Pure language instruction data.}
We also include an off-domain auxiliary source from the NVIDIA Nemotron instruction-following dataset, \textit{nvidia/Nemotron-SFT-Instruction-Following-Chat-v2}. This dataset is text-only and targets open-ended chat and instruction-following supervision, rather than visual grounding or robot control. We include it as a control-like auxiliary signal: if pure language supervision improves downstream VLA performance, then the benefit of auxiliary co-training cannot be explained solely by direct transfer from vision-language or robotics-related tasks. Instead, such a gain supports our representation-level interpretation: non-action supervision can help preserve the backbone's foundation-model operating regime and diversify feature updates during VLA pre-training.

\begin{figure}[h]
\centering
\includegraphics[width=0.95\linewidth]{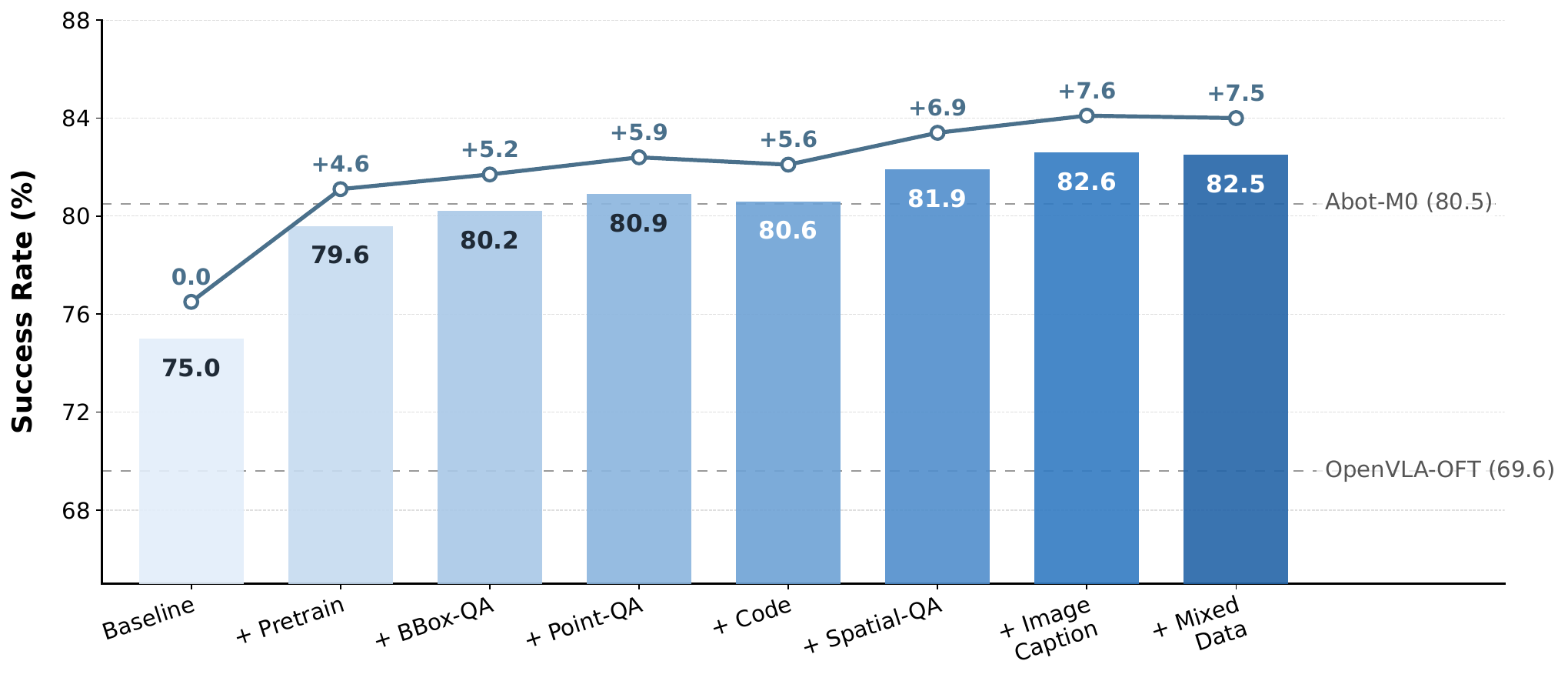}
\caption{\textbf{Effect of auxiliary co-training data.}
We compare different non-action supervision sources under the same robot data, model, and pre-training budget. Caption data gives the strongest single-source gain, while grounding, spatial-QA, and pure language instruction data also improve over robot-only training.}
\label{fig:aux_data_cotrain}
\end{figure}

\subsection{Effect of Different Auxiliary Co-training Data}
\label{sec:vlm_data_ablation}

We compare different auxiliary supervision types while keeping the robot data, backbone, training recipe, and total pre-training budget fixed. Fig.~\ref{fig:aux_data_cotrain} reports the LIBERO-Plus results. Relative to the robot-only baseline, all auxiliary sources improve downstream performance, indicating that non-action supervision helps produce a stronger VLA-oriented backbone.

Image-caption data gives the strongest single-source gain, improving the baseline from 75.0 to 82.6. This is consistent with its role as the main representation-preservation signal: detailed captions are close to the original VLM pre-training distribution and provide dense semantic supervision over objects, attributes, relations, and scene context. Spatial-QA, Point-QA, and BBox-QA also improve performance, suggesting that spatial reasoning and localized grounding provide useful complementary constraints on the learned representation.

Pure language instruction data also improves over the robot-only baseline. Since this source is text-only and not directly related to robot perception, visual grounding, or control, its gain is especially informative. It suggests that auxiliary co-training does not help only through direct transfer from VLM or robotics-related tasks. Instead, non-action supervision can help preserve the backbone's foundation-model representation and make feature updates more diverse than action-only training.

The mixed-data setting reaches 82.5, slightly below caption-only co-training. Since the total number of pre-training steps is fixed, mixing multiple auxiliary sources reduces the sampling frequency of image-caption data, which is the strongest single source in this ablation. This likely pulls the mixed result slightly below caption-only co-training, rather than indicating that data mixing itself is harmful.

Overall, this ablation supports the representation-preservation role of auxiliary co-training. Caption data provides the strongest single anchor, while grounding, spatial, and pure language instruction data provide additional gains. These results justify using auxiliary data not as generic extra supervision, but as a targeted mechanism for preserving and enriching the backbone representation during VLA pre-training.
\section{Representation Effects of Head-Diverse Pre-training}
\label{sec:appendix_heads}

\subsection{Head-Diverse Pre-training Prevents Decoder Lock-in}
\label{sec:decoder_lockin}

The pilot study suggests that single-head continuous pre-training can reduce the reusability of the backbone with a different downstream action head. We refer to this tendency as \emph{decoder lock-in}: the backbone may organize action-relevant features in a form that is effective for the pre-training decoder, but less accessible to alternative decoders introduced during fine-tuning. This is a concern for building a reusable VLA-oriented backbone, since downstream users may adopt different action heads depending on the embodiment, control frequency, action horizon, or deployment constraints.

Table~\ref{tab:unseen_head_probe} examines this issue using PI as the downstream fine-tuning head. Without additional continuous pre-training, PI fine-tuning reaches 60.5. After OFT-only pre-training, PI fine-tuning drops to 55.1, suggesting that the OFT pre-trained representation is not necessarily more compatible with the PI action head. In contrast, when the backbone is pre-trained with two continuous heads, OFT and GR00T, PI fine-tuning improves to 63.1, even though PI is not used during pre-training. The improvement is modest, but the comparison between OFT-only and OFT+GR00T is informative: adding a second pre-training head changes the unseen PI fine-tuning result from below the from scratch baseline to slightly above it.

When PI is also included during pre-training, PI fine-tuning reaches 77.0. This result should be interpreted separately, since the downstream head is now directly supervised during pre-training. It confirms that direct head exposure provides a strong compatibility benefit, but the more relevant comparison for decoder lock-in is between OFT-only and OFT+GR00T. Overall, this pilot study suggests that head diversity can reduce over-specialization to a single decoder and improve the accessibility of the learned representation to an unseen downstream head.

\begin{table}[t]
\centering
\small
\caption{\textbf{Head diversity reduces decoder lock-in.}
PI fine-tuning degrades after OFT-only pre-training, but improves after OFT+GR00T pre-training even though PI is excluded during pre-training. Light blue highlights the key unseen-head comparison; dark blue marks the full configuration with direct PI exposure.}
\label{tab:unseen_head_probe}
\begin{tabular}{lccc}
\toprule
\textbf{Pre-training heads}
& \textbf{Seen PI in pre-training}
& \textbf{PI FT}
& \textbf{$\Delta$ vs. scratch} \\
\midrule
None & -- & 60.5 & -- \\
OFT & No & 55.1 & $-$5.4 \\
\rowcolor{tableSecond}
OFT + GR00T & No & \secondcell{63.1} & $+$2.6 \\
\rowcolor{tableBest}
OFT + PI + GR00T & Yes & \bestcell{77.0} & $+$16.5 \\
\bottomrule
\end{tabular}
\end{table}

\subsection{Does Head Diversity Also Improve Same-head Adaptation?}
\label{sec:same_head_adaptation}

The main motivation for head-diverse pre-training is to improve compatibility across action heads. We further examine whether this comes at the cost of same-head adaptation, where the downstream fine-tuning head matches one of the heads used during pre-training. Ideally, a more head-agnostic representation should remain competitive when the same head is reused.

Table~\ref{tab:same_head_rep_quality} compares three settings: fine-tuning without additional continuous pre-training, fine-tuning from a matched single-head pre-trained backbone, and fine-tuning from the full head-diverse pre-trained backbone. Head-diverse pre-training improves over the corresponding single-head baselines for OFT, PI, and GR00T, with gains of 1.7, 1.6, and 4.3 points, respectively. These gains are relatively moderate compared with the direct benefit of pre-training itself, but they indicate that multi-head supervision does not harm same-head adaptation in our setting.

This result is consistent with the representation-level view in Sec.~\ref{sec:method-align}. When the backbone is trained with multiple decoders, it must expose action information in a form that can be read by several action parameterizations. This reduces over-specialization to any single decoder and can produce action features that are more useful even when the same head is reused.

\begin{table}[t]
\centering
\small
\caption{\textbf{Same-head adaptation under head-diverse pre-training.}
Head-diverse pre-training improves downstream performance even when the fine-tuning head matches one of the pre-training heads. Within each row, blue and lighter-blue cells mark the best and second-best results.}
\label{tab:same_head_rep_quality}
\begin{tabular}{lcccc}
\toprule
\textbf{Fine-tune head}
& \textbf{From scratch}
& \textbf{Single-head pre-train}
& \textbf{Head-diverse pre-train}
& \textbf{$\Delta$ vs. single} \\
\midrule
OFT   & 61.7 & \secondcell{78.8} & \bestcell{80.5} & $+$1.7 \\
PI    & 60.5 & \secondcell{75.4} & \bestcell{77.0} & $+$1.6 \\
GR00T & 51.2 & \secondcell{71.7} & \bestcell{76.0} & $+$4.3 \\
\bottomrule
\end{tabular}
\end{table}

\mypara{Summary.}
These diagnostics support the role of head diversity as a representation-compatibility mechanism. Single-head continuous pre-training can lock the backbone into one decoder geometry, making alternative downstream heads difficult to optimize. Adding multiple heads during pre-training reduces this lock-in by requiring the shared latent representation to support more than one action head. The same representation also improves same-head adaptation in our experiments, but the central objective is broader: to obtain a reusable VLA backbone that can support diverse downstream action heads.

\section{Unified Joint Space and Wrap-Aware Loss}
\label{sec:wrap_loss_detail}

\subsection{Unified joint space by wrapping.}
Absolute joint angles are naturally defined on a periodic space rather than on the Euclidean line.
In particular, each revolute joint angle should be represented within the canonical range $[-\pi,\pi]$.
Angles outside this interval do not correspond to new physical joint states; they are merely alternative parameterizations of equivalent poses.
For example, $\pi+\epsilon$ and $-\pi+\epsilon$ describe nearly identical joint configurations, but their raw numerical values differ by almost $2\pi$.
If such equivalent angles are kept as different supervision targets, the model is forced to produce different outputs for the same underlying physical state, making it harder to learn a consistent action representation.
Therefore, before training, we wrap all absolute joint-angle dimensions into $[-\pi,\pi]$:
\begin{equation}
a_{\mathrm{wrap}}
=
\big(a+\pi\big)\bmod 2\pi-\pi .
\label{eq:data_wrap}
\end{equation}
This data-side wrapping constructs a unified joint space, where equivalent joint states are mapped to a consistent numerical representation.

\mypara{Wrap-aware loss.}
Although data-side wrapping ensures that target actions lie in the canonical range, a standard regression loss can still produce an incorrect distance measure near the wrap boundary.
For instance, when the target angle is close to $-\pi$ and the prediction is close to $+\pi$, the two angles are physically close, but the naive residual $\hat{a}-a$ is close to $2\pi$.
To correct this artifact, we further wrap the residual between the predicted and target actions before computing the regression loss.
Given a predicted action $\hat{a}$ and a target action $a$ on a revolute dimension, we define
\begin{equation}
\delta_{\mathrm{wrap}}
=
\big((\hat{a}-a)+\pi\big)\bmod 2\pi-\pi,
\label{eq:residual_wrap}
\end{equation}
and compute the wrap-aware loss with a standard L1 penalty:
\begin{equation}
\mathcal{L}_{\mathrm{wrap}}
=
\left|\delta_{\mathrm{wrap}}\right|.
\label{eq:wrap_loss}
\end{equation}

We add this term to the original training objective of each action head.
Specifically, for a head $h$, we denote its original action-learning objective as $\mathcal{L}^{(h)}_{\mathrm{action}}$.
This objective is the standard loss used by the corresponding head architecture, such as direct action regression for OFT, diffusion training loss for GR00T, or flow-matching loss for PI.
The final objective for head $h$ is therefore
\begin{equation}
\mathcal{L}^{(h)}_{\mathrm{total}}
=
\mathcal{L}^{(h)}_{\mathrm{action}}
+
\mathcal{L}^{(h)}_{\mathrm{wrap}},
\label{eq:total_wrap_loss}
\end{equation}
where $\mathcal{L}^{(h)}_{\mathrm{wrap}}$ is computed on the final action prediction of head $h$.
For OFT, this corresponds to its direct regression output.
For GR00T and PI, this corresponds to the final generated action sample after the denoising or flow-based generation process, rather than the intermediate noise prediction or velocity target.
The wrap-aware term is applied only to absolute joint-angle dimensions.
Non-periodic dimensions, such as gripper commands and delta end-effector translations, are excluded.

\subsection{Experiments.}
We ablate the effect of data-side joint-angle wrapping and loss-side residual wrapping on RoboTwin under the base setting.
Specifically, we train each model using $50$ tasks with $50$ trajectories per task, and evaluate the trained policy on the Clean evaluation split.
For a fair comparison, all variants use the same pre-training and fine-tuning protocol, and only differ in whether unified joint-space wrapping and wrap-aware loss are enabled.
As shown in Table~\ref{tab:wrap_ablation}, the baseline that directly regresses raw joint angles achieves $75.5\%$ success rate.
Introducing the unified joint space improves the performance to $78.6\%$, suggesting that data-side wrapping reduces ambiguity in the action targets.
Adding the wrap-aware loss further improves the success rate to $80.5\%$, showing that residual wrapping provides a more appropriate distance measure near the angular boundary.

\begin{table}[t]
\centering
\small
\setlength{\tabcolsep}{6pt}
\caption{\textbf{Ablation of unified joint space and wrap-aware loss on RoboTwin.}
All models are trained under the base setting using $50$ tasks with $50$ trajectories per task, and evaluated on the Clean split.
The pre-training and fine-tuning protocols are kept identical across all variants. Dark blue highlights the full design; light blue highlights the strongest partial variant.}
\label{tab:wrap_ablation}
\begin{tabular}{lccc}
\toprule
\textbf{Setting}
& \textbf{Unified Joint Space}
& \textbf{Wrap Loss}
& \textbf{Robotwin} \\
\midrule
Baseline
&  
&  
& 75.5 \\
\rowcolor{tableSecond}
Unified joint space
& $\checkmark$
&  
& \secondcell{78.6} \\
\rowcolor{tableBest}
Unified joint space + Wrap-aware loss
& $\checkmark$
& $\checkmark$
& \bestcell{80.5} \\
\bottomrule
\end{tabular}
\end{table}

\section{Adding Heterogeneous UMI Data to Continued Pre-training}
\label{sec:data_scaling_realomin}

Our main results use a fixed open-source pre-training mixture. Here we check whether the recipe can also make use of trajectories collected under a different embodiment and a different collection paradigm. We add 20K trajectories from 10Kh-RealOmin-OpenData~\citep{genrobot2026realomin}, a corpus of real-world manipulation trajectories collected through a UMI-style handheld interface.

\mypara{Setup.}
We select 20K trajectories, filtering primarily on trajectory continuity and replayability, and convert their actions into delta end-effector representations so that they are compatible with our partially unified action layout. All other settings are unchanged, including the model architecture, the rest of the pre-training mixture, the optimization budget, and the downstream fine-tuning protocol.

\begin{table}[t]
\centering
\small
\caption{\textbf{Adding heterogeneous UMI trajectories to continued pre-training.}
Adding 20K trajectories from 10Kh-RealOmin-OpenData~\citep{genrobot2026realomin}, collected with a different embodiment and a UMI-style interface, improves downstream LIBERO-Plus performance. All other training and downstream fine-tuning settings are unchanged.}
\label{tab:realomin_scaling}
\begin{tabular}{lc}
\toprule
\textbf{Continued pre-training data} & \textbf{LIBERO-Plus} \\
\midrule
Original VLAct mixture & 82.6 \\
\rowcolor{tableBest}
Original VLAct mixture + 20K RealOmin trajectories & \bestcell{83.7} \\
\bottomrule
\end{tabular}
\end{table}

\mypara{Result.}
As shown in Table~\ref{tab:realomin_scaling}, adding the 20K UMI trajectories improves LIBERO-Plus from 82.6 to 83.7, a gain of 1.1 points. Since these trajectories differ from the original mixture in both embodiment and collection paradigm, and received only lightweight filtering without task-specific curation, we read this as evidence that the recipe can absorb heterogeneous data rather than being disrupted by it.

\section{Details of Data Cleaning}
\label{sec:data_clean_detail}

\paragraph{Robot Platforms and Action Spaces.}
We train our models on two robot platforms: a Franka single-arm platform and an AgileX dual-arm platform.
For the AgileX dual-arm setting, we train on AgileX data from InternData-A1 and RoboCoin, and evaluate the trained models on RoboTwin.
The AgileX action space is 14-dimensional, consisting of 12 absolute joint-angle dimensions for the two arms and 2 gripper dimensions.
For the Franka single-arm setting, we train on DROID and MolmoAct, where DROID includes both v1.0.0 and v1.0.1, and evaluate on LIBERO-Plus and VLA-Arena.
The Franka action space is 7-dimensional, consisting of 6-dimensional delta end-effector actions and 1 gripper dimension.

\paragraph{Unified Cleaning Pipeline.}
The training data are aggregated from multiple robot datasets collected with different platforms, annotation conventions, control interfaces, action spaces, and preprocessing pipelines.
As a result, the raw data contain several sources of heterogeneity and noise.
First, some trajectories have invalid or uninformative task names, which cannot provide meaningful language supervision for instruction-conditioned policy learning.
Second, some action sequences contain abnormal values with unrealistically large magnitudes, which are likely caused by data collection failures, logging errors, or preprocessing artifacts.
Third, even when the trajectories are valid, different datasets may use different action scales, control frequencies, and gripper conventions.
Without proper alignment, the model may learn dataset-specific action patterns rather than a unified action representation that generalizes across datasets.

To address these issues, we apply a unified data cleaning pipeline before training.
The pipeline aims to remove unreliable supervision signals while preserving useful trajectories as much as possible.
It mainly consists of two stages.
First, we remove trajectories or samples with invalid task names, including \texttt{unknown\_task}, \texttt{n/a}, \texttt{none}, \texttt{no action}, whitespace-only task names, and empty task names.
Such annotations do not provide meaningful language supervision and may introduce ambiguity during instruction-conditioned policy learning.
Second, we filter action chunks that contain an excessive number of abnormal actions with unusually large magnitudes.
This step removes severe action artifacts while avoiding overly aggressive trajectory-level filtering.

\paragraph{Delta End-Effector Action Alignment.}
For delta end-effector actions, different datasets may be collected at different frame rates, while the physical robot motion speed is generally comparable across datasets.
To reduce discrepancies caused by different FPS values, we rescale each delta action by the corresponding FPS and convert it to a unified per-second representation.
An action step is marked as invalid if its absolute translational velocity exceeds 0.5 or its absolute rotational velocity exceeds 1.0.

Instead of discarding an entire trajectory whenever abnormal actions appear, we adopt a step-level masking strategy.
Specifically, invalid action steps are masked out during training, so that a small number of noisy actions does not remove an otherwise useful trajectory.
For each action chunk $\mathcal{C}$, we compute the invalid-step ratio:
\[
r_{\mathcal{C}} = \frac{1}{|\mathcal{C}|}\sum_{t \in \mathcal{C}} \mathbb{I}\left[\text{step } t \text{ is invalid}\right].
\]
We discard the whole chunk only when $r_{\mathcal{C}} > 0.5$, since such chunks are dominated by unreliable supervision signals.
After filtering, we normalize the translational and rotational components by 0.5 and 1.0, respectively.
This maps delta end-effector actions from different datasets into a shared and numerically stable action space.

\paragraph{Absolute Joint-Angle Action Alignment.}
For absolute joint-angle actions, we first remove values outside the range $[-2\pi, 2\pi]$, as such values are likely caused by data collection errors or preprocessing artifacts.
For the remaining valid actions, we wrap each joint angle into the range $[-\pi, \pi]$:
\[
\theta_{\mathrm{wrap}} = ((\theta + \pi) \bmod 2\pi) - \pi .
\]
This wrapping operation resolves equivalent angular representations and ensures that all joint-angle actions follow a consistent convention.
After this step, joint-angle actions from different datasets naturally lie in the same angular space, and therefore we do not apply additional normalization to them.

\paragraph{Gripper Action Normalization.}
For gripper actions on both robot platforms, we first remove extreme outliers and then apply min-max normalization separately for each dataset.
This maps gripper values to continuous values in $[0, 1]$, providing a consistent representation across datasets despite differences in gripper hardware, control interfaces, and recording conventions.

\section{Details of Action Heads}
\label{app:action-heads}

\subsection{Representative Action-Head Families}
\label{sec:head_detail_raw}

\begin{figure}[t]
    \centering
    \includegraphics[width=1\linewidth]{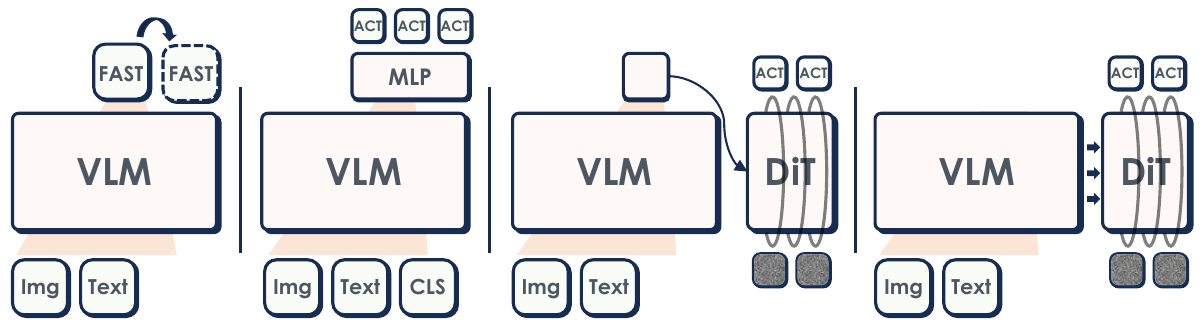}
    \caption{
    Comparison of representative VLA action heads studied in this work.
    FAST preserves the VLM's autoregressive interface by discretizing action
    chunks into action tokens. OFT attaches a lightweight MLP head to VLM
    action features and predicts continuous actions in parallel. PI and GR00T
    use a separate DiT-style motor module conditioned on VLM features to
    generate continuous action chunks through flow matching, with GR00T using
    a more explicitly decoupled dual-system design.
    }
    \label{fig:action-head-comparison}
\end{figure}

We briefly summarize the four representative action-head families used in our
study.  They share the same high-level role: mapping VLM representations to
robot action chunks.  Their main difference lies in the action interface:
FAST uses discrete autoregressive action tokens, OFT uses direct continuous
regression, and PI/GR00T use flow-matching-based continuous action generation.

\mypara{FAST: autoregressive discrete action tokens.}
FAST converts a continuous action chunk into a sequence of discrete action
tokens.  Let
\[
    z_{1:M} = \mathrm{Tok}_{\mathrm{FAST}}(A_t)
\]
denote the token sequence obtained from the continuous action chunk.  The VLA
model is trained with the standard next-token prediction objective
\[
    \mathcal{L}_{\mathrm{FAST}}
    =
    - \sum_{m=1}^{M}
    \log p_\theta(z_m \mid H_t, z_{<m}).
\]
At inference time, the model autoregressively generates action tokens
$\widehat{z}_{1:M}$ and maps them back to continuous actions through the
inverse tokenizer,
\[
    \widehat{A}_t
    =
    \mathrm{Tok}_{\mathrm{FAST}}^{-1}(\widehat{z}_{1:M}).
\]
FAST therefore preserves the native language-modeling form of the VLM: action
prediction is reduced to token prediction.  Compared with direct per-dimension
action binning, FAST-style tokenization compresses the action trajectory before
discretization, often by representing action chunks in a frequency-domain or
otherwise compact basis.  Its advantage is architectural compatibility with
autoregressive VLMs; its cost is that continuous control is mediated through a
discrete tokenizer and decoded sequentially.

\mypara{OFT: parallel continuous regression.}
OFT removes the discrete tokenizer and attaches a lightweight continuous
regression head to the VLM.  The model appends a fixed set of action query
tokens to the backbone sequence.  If $h^{\mathrm{act}}_{t,k}$ denotes the
hidden state corresponding to the $k$-th action query, then a small MLP predicts
the corresponding action vector:
\[
    \widehat{a}_{t+k}
    =
    g_\phi(h^{\mathrm{act}}_{t,k}),
    \qquad
    k=0,\ldots,K-1.
\]
The head is trained with a pointwise regression loss, typically
\[
    \mathcal{L}_{\mathrm{OFT}}
    =
    \frac{1}{K d_a}
    \sum_{k=0}^{K-1}
    \left\|
        g_\phi(h^{\mathrm{act}}_{t,k}) - a_{t+k}
    \right\|_1 .
\]
Unlike FAST, OFT predicts the entire action chunk in one forward pass.  This
makes OFT a simple and efficient probe of whether the backbone representation
linearly exposes the information needed for continuous control.  The tradeoff
is that the output is a point estimate rather than an explicit multimodal
action distribution.

\mypara{PI: flow-matching action expert.}
PI-style heads model continuous action chunks through conditional flow
matching.  Instead of directly predicting $A_t$, the head learns a vector field
that transports Gaussian noise into the demonstrated action chunk conditional
on the backbone representation.  Let $\epsilon \sim \mathcal{N}(0,I)$ and
sample a flow time $\tau \in [0,1]$.  With a linear probability path,
\[
    A_t^\tau = (1-\tau)\epsilon + \tau A_t,
\]
the target velocity is
\[
    u_t = A_t - \epsilon .
\]
The action expert $v_\phi$ receives the noised action chunk, the flow time, and
the backbone representation, and is trained by
\[
    \mathcal{L}_{\mathrm{PI}}
    =
    \mathbb{E}_{A_t,\epsilon,\tau}
    \left[
        \left\|
            v_\phi(A_t^\tau,\tau;H_t) - (A_t-\epsilon)
        \right\|_2^2
    \right].
\]
At inference time, the head starts from Gaussian noise
$A_t^0 \sim \mathcal{N}(0,I)$ and integrates the learned vector field:
\[
    A_t^{n+1}
    =
    A_t^n
    +
    \Delta \,
    v_\phi(A_t^n,\tau_n;H_t),
    \qquad
    n=0,\ldots,N-1,
\]
returning $\widehat{A}_t = A_t^N$.  Compared with OFT, PI introduces an
iterative generative decoder.  This is more expensive than one-shot regression,
but it can represent richer continuous action distributions and separates
semantic conditioning from motor generation more explicitly.

\mypara{GR00T: dual-system flow-matching motor module.}
GR00T also uses flow matching, but places it inside a more explicitly separated
dual-system architecture.  The vision-language backbone acts as a slow
semantic reasoning module, while a DiT-style action module acts as a fast
motor-generation module.  The backbone encodes images and language into
vision-language tokens.  These tokens, together with proprioceptive state
encodings and noised action encodings, condition a diffusion-transformer action
module.

Formally, the training objective has the same flow-matching form:
\[
    \mathcal{L}_{\mathrm{GR00T}}
    =
    \mathbb{E}_{A_t,\epsilon,\tau}
    \left[
        \left\|
            v_\phi(A_t^\tau,\tau;H_t,s_t,e) - (A_t-\epsilon)
        \right\|_2^2
    \right],
\]
where $s_t$ denotes the robot state and $e$ denotes an embodiment identifier or
embodiment-specific adapter.  The key distinction from PI is architectural:
state and action vectors are embedded into a separate motor module, processed
by a DiT-style transformer, and decoded back into the robot's native action
space.  Cross-attention to the VLM tokens keeps the motor module conditioned on
the scene and instruction while preserving a specialized action-generation
pathway.

\subsection{Unified Action Representation}
\label{sec:unified-output-layout}

Although FAST, OFT, PI, and GR00T use different decoding mechanisms, they all
ultimately supervise a future action chunk.  In our multi-embodiment
pre-training, we therefore map each dataset's native action format into a
shared canonical output space before applying the corresponding head-specific
objective.

\mypara{Design}
For the continuous heads used in our main pre-training recipe, the action head
emits a fixed $20$-dimensional vector at each action step,
\[
    \widehat{a}_{t+k} \in \mathbb{R}^{20},
    \qquad k=0,\ldots,K-1 .
\]
The dimensions are assigned by embodiment role:
\begin{itemize}

    \item Dimensions $1$--$6$: the left $6$-DoF arm of bimanual embodiments
    such as AgileX, represented as absolute joint angles.

    \item Dimensions $7$--$12$: the right $6$-DoF arm of bimanual embodiments
    such as AgileX, represented as absolute joint angles.

    \item Dimensions $13$--$18$: the $6$-DoF arm of single-arm embodiments
    such as Franka, represented as delta end-effector pose.

    \item Dimension $19$: the shared gripper coordinate, used by Franka's
    single gripper and AgileX's left gripper.

    \item Dimension $20$: the right gripper coordinate of bimanual
    embodiments.
\end{itemize}

This layout preserves each corpus's native low-level action convention: Franka
actions use delta end-effector pose in DROID and MolmoAct, while AgileX actions
use absolute joint angles in InternData-A1 and RoboCoin. Thus, the unified
head design does not force all embodiments into the same controller parameterization.
It only aligns output coordinates at the level of embodiment role, allowing the
same FAST/OFT/PI/GR00T head families to be instantiated on a common cross-embodiment action representation.

\mypara{Experiments}
We evaluate the effect of the unified action representation design under three settings:
separate embodiment-specific heads, a unified head without action space
alignment, and a unified head with unified action representation. Experiments are
conducted on RoboTwin and LIBERO-Plus. As shown in
Table~\ref{tab:unified-output-layout}, simply sharing the action head already improves
performance over separate heads, while the fully unified action representation
achieves the best results on both benchmarks. This suggests that aligning
outputs by embodiment role provide a more reusable and transferable action
representation across robots.

\begin{table}[t]
\centering
\small
\setlength{\tabcolsep}{6pt}
\caption{\textbf{Effect of unified action representation design.} We compare separate heads,
a unified head, and the full unified action representation on RoboTwin and
LIBERO-Plus. Dark blue highlights the full design; light blue highlights the unified-head intermediate.}
\label{tab:unified-output-layout}
\begin{tabular}{lcc}
\toprule
\textbf{Setting} & \textbf{RoboTwin} & \textbf{LIBERO-Plus} \\
\midrule
Separate heads & 78.5 & 81.1 \\
\rowcolor{tableSecond}
Unified head & \secondcell{79.5} & \secondcell{81.4} \\
\rowcolor{tableBest}
Unified action representation & \bestcell{80.5} & \bestcell{82.6} \\
\bottomrule
\end{tabular}
\end{table}

\section{Details of Real-world Experiments}
\label{app:Real-world-exp}
\begin{figure}[t]
    \centering
    \includegraphics[width=1.\linewidth]{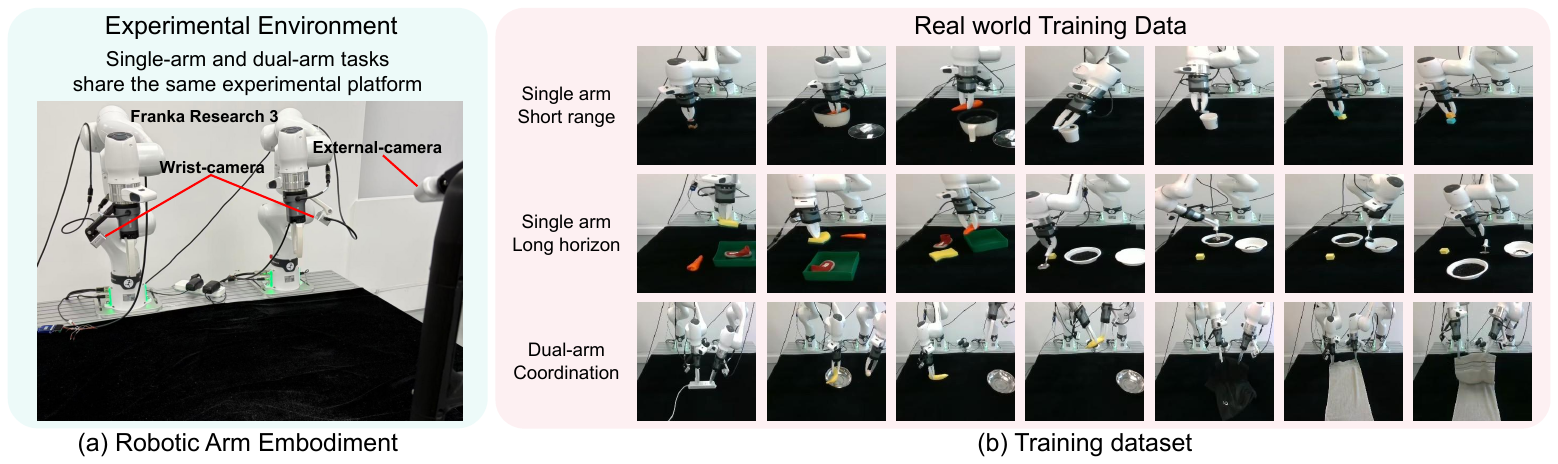}
    \caption{Overviews of our robot setting and real-world training dataset. (a) Robot setting for the experiments. (b) We collect real-world robot trajectories for eleven task suites.}
    \label{fig:app-real-setting}
\end{figure}

\begin{figure}[t]
    \centering
    \includegraphics[width=1.\linewidth]{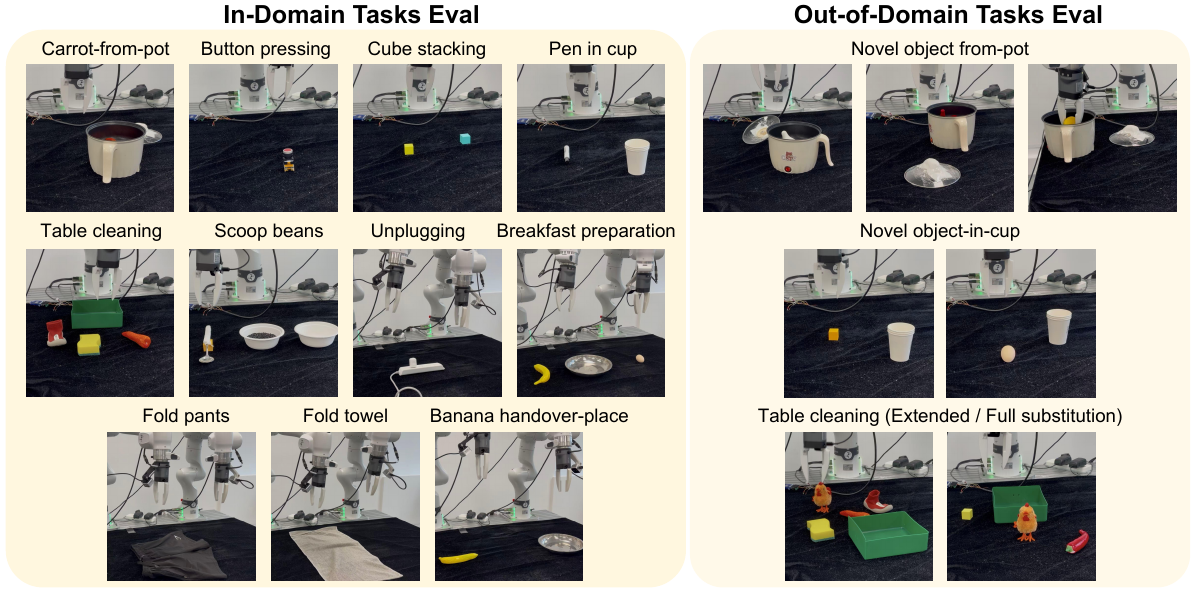}
    \caption{Schematic illustrations of real-robot in-domain and out-of-domain evaluation tasks.}
    \label{fig:app-real-tasks}
\end{figure}

\subsection{Hardware and observation.}
As shown in Fig.~\ref{fig:app-real-setting}, our physical evaluation platform uses stationary, table-mounted Franka Research 3 7-DoF arms (one for single-arm setups, two for dual-arm tasks). The visual system is a multi-view configuration: a single Intel RealSense D435 is fixed as an external camera providing a third-person perspective, and each robot arm carries a wrist-mounted Intel RealSense D405 for precise first-person observations. In line with the input specification of our representation-centric pre-training recipe, all images from external and wrist cameras are resized to $224\!\times\!224$ before being processed by the model.

\subsection{Expert Data Collection and Fine-tuning Protocol}
\label{app:real-world-protocol}

\mypara{Expert data collection.}
All real-world demonstrations are collected by human teleoperation using a GELLO~\citep{wu2024gello} interface. During collection, we randomize the initial placements of the manipulated objects and the initial configuration of the robot arms across episodes, so that the demonstrations cover a range of starting states rather than repeating a single scripted trajectory. Fig.~\ref{fig:app-real-setting}(b) shows the eleven task suites for which we collect data.

\mypara{Fine-tuning data.}
Each single-arm task uses 50 demonstrations, and each dual-arm task uses 100 demonstrations, reflecting the higher coordination complexity of bimanual manipulation. VLAct and the Qwen3VL-4B-OFT baseline are fine-tuned on \emph{exactly the same} demonstrations, under the same action head, optimizer, and training budget, so that the comparison isolates the pretrained backbone rather than the downstream data or recipe.

\mypara{Evaluation protocol.}
For each task, both models are evaluated from the same 10 fixed robot and object initial configurations, held constant across models and across runs. These are the 10 rollouts per task used in the success-rate computation described below. Because the initial states are fixed and shared, differences in success rate reflect policy behavior rather than variation in starting conditions.

\subsection{Detailed Real-World Results and Failure Analysis}
\mypara{Success Rate Calculation Criterion.}
For real-machine experiments, each task is rolled out for 10 trajectories to calculate the task success rate. The scoring rules vary for different task types. Specifically, for single-arm short-range tasks and dual-arm collaborative tasks, a score of 1 is awarded for a successful task completion and 0 for a failure. For single-arm long-range tasks, the scoring mechanism is based on the number of completed task steps. Taking the table cleaning task as an example, the entire task consists of three independent steps, and each successfully completed step contributes 0.33 points to the total task score.

\mypara{Single-arm short-horizon tasks.}
VLAct achieves an average success rate of 92.5\% on in-domain single-arm
short-horizon tasks, compared with 77.5\% for Qwen3VL-4B-OFT without
pre-training. Both models perform well on simpler tasks such as
\textit{carrot-from-pot}, where both reach 100\% SR, and \textit{button
pressing}, where VLAct reaches 100\% SR and the baseline reaches 90\% SR.
However, the gap becomes larger when the task requires more precise spatial
localization. On \textit{cube stacking}, VLAct reaches 90\% SR while the
baseline drops to 60\%. We observe that the baseline often freezes during
inference when the cube starts from positions not well covered by the training
distribution. On \textit{pen-in-cup}, VLAct reaches 80\% SR compared with 60\%
for the baseline. VLAct can localize the cup even when it is placed near the rear
of the table, while the baseline often accumulates errors after failed grasping
attempts and enters unstable recovery behaviors.

\mypara{Short-horizon OOD generalization.}
VLAct also generalizes better to unseen objects. For \textit{novel object
from-pot}, where the carrot is replaced by egg, pepper, or garlic, VLAct obtains
90.0\% average SR, while the baseline reaches 73.3\%. For \textit{novel object
in-cup}, where the target object is replaced by cube or egg, VLAct again reaches
90.0\% average SR, compared with 65.0\% for the baseline. These results indicate
that the pre-trained backbone provides stronger visual-semantic grounding and
more stable localization under object-level distribution shift.

\mypara{Single-arm long-horizon tasks.}
For long-horizon tasks, VLAct shows a clearer advantage in preserving task logic
over multiple manipulation stages. On \textit{table cleaning}, VLAct obtains a
weighted success rate 86.6\%, compared with 73.3\% for the baseline. On
\textit{scooping beans}, VLAct reaches 80.0\%, whereas the baseline only reaches
33.3\%. In this task, VLAct usually completes the intended sequence: grasping the
spoon, scooping a bean, and pouring it into the target bowl. By contrast, the
baseline frequently skips the scooping phase and directly performs a pouring
motion with an empty spoon. In cluttered table-cleaning scenes, the baseline also
tends to overlook small objects, while VLAct completes the sequence more
fluently.

\mypara{Long-horizon OOD generalization.}
We further evaluate long-horizon robustness under two harder OOD settings. In
the extended-sequence setting, a novel toy chicken is added to the original table
cleaning task. VLAct achieves a weighted success rate of 82.5\%, nearly doubling
the baseline score of 47.5\%. In the full-substitution setting, all objects are
replaced by unseen categories, including cube, toy chicken, and pepper. VLAct
maintains a score of 83.3\%, while the baseline drops to 46.6\%. The baseline often
misses objects or fails to recover from grasping errors, whereas VLAct maintains
the task sequence under both object replacement and sequence extension.

\mypara{Dual-arm coordination.}
Across dual-arm coordination tasks, VLAct achieves 72.0\% average SR,
outperforming the baseline's 44.0\% SR. On \textit{breakfast preparation},
VLAct reaches 90.0\% SR compared with 70.0\% for the baseline. On
\textit{unplugging}, VLAct reaches 80.0\% SR compared with 60.0\% for the baseline,
showing better coordination in the hold-and-pull interaction. The largest gaps
appear in deformable-object manipulation. On \textit{fold pants}, VLAct achieves
70.0\% SR, outperforming the baseline by 30.0\%. The baseline often
selects inaccurate grasping points on the fabric, leading to gripper-table
collisions or failed lifting. On \textit{fold towel}, VLAct reaches 50.0\% SR,
while the baseline reaches only 20.0\%. The baseline frequently suffers from 
velocity mismatch between the two arms, causing the towel to slip from one
gripper. VLAct produces more synchronized dual-arm motion and more stable
contact with deformable objects.

\subsection{Detailed Prompts for Each Real-World Robotic Arm Task}
\textbf{Single-arm (Short-horizon)}
\begin{itemize}
    \item \textbf{Carrot-from-pot:} Pick the carrot out of the pot and place it on the table.
    \item \textbf{Button pressing:} Press the red button.
    \item \textbf{Cube stacking:} Pick up one cube and stack it on top of another cube.
    \item \textbf{Pen in cup:} Pick up the pen and place it into the paper cup.
\end{itemize}

\textbf{Single-arm (Long-horizon)}
\begin{itemize}
    \item \textbf{Table cleaning:} Pick up the small objects on the table one by one and place them into the green box.
    \item \textbf{Scoop beans:} First pick up the spoon, then use it to scoop a bean from one bowl, and finally pour the bean into another bowl.
\end{itemize}

\textbf{Dual-arm Coordination}
\begin{itemize}
    \item \textbf{Unplugging:} One robotic arm holds the socket in place, while the other pulls the plug out and places it on the table.
    \item \textbf{Breakfast preparation:} One robotic arm picks up the banana, while the other picks up the egg; both are then placed on the plate.
    \item \textbf{Banana handover-place:} One robotic arm picks up the banana and hands it to the other arm; the second arm receives it and places it on the plate.
    \item \textbf{Fold pants:} Two robotic arms fold a pair of pants together.
    \item \textbf{Fold towel:} Two robotic arms fold a towel together.
\end{itemize}

\textbf{Out-of-Domain (OOD)}
\begin{itemize}
    \item \textbf{Novel object from-pot:} 
    Pick the garlic / pepper / egg out of the pot and place it on the table.
    \item \textbf{Novel object in-cup:} 
    Pick up the cube / egg and place it into the paper cup.
    \item \textbf{Table cleaning (extended / full substitution):} 
    Pick up the small objects on the table one by one and place them into the green box.
\end{itemize}

\subsection{Visualization}
\begin{figure}[p]
\centering
\includegraphics[width=0.95\linewidth,height=0.88\textheight,keepaspectratio]{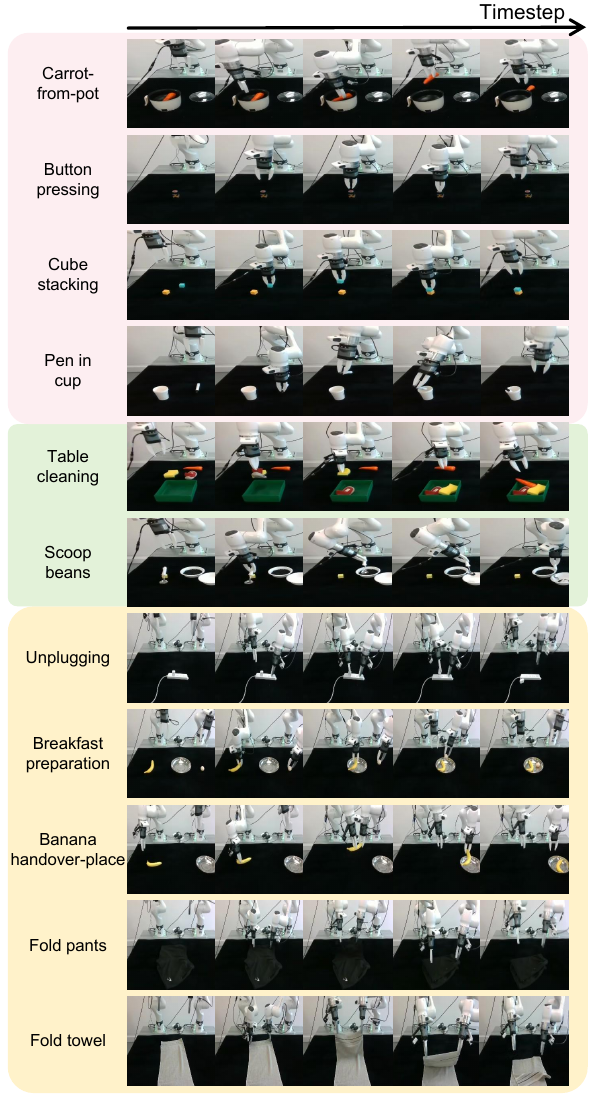}
\caption{\textbf{Detailed Visualization of Training Data for Each Real-World Robotic Arm Task.}}
\label{fig:real_world_train_task_visual}
\end{figure}

\mypara{Detailed Visualization of Training Data for Each Real-World Robotic Arm Task. }
Fig.~\ref{fig:real_world_train_task_visual} shows the visualization of training data for each real-world robotic arm task.

\begin{figure}[h]
\centering
\includegraphics[width=0.95\linewidth]{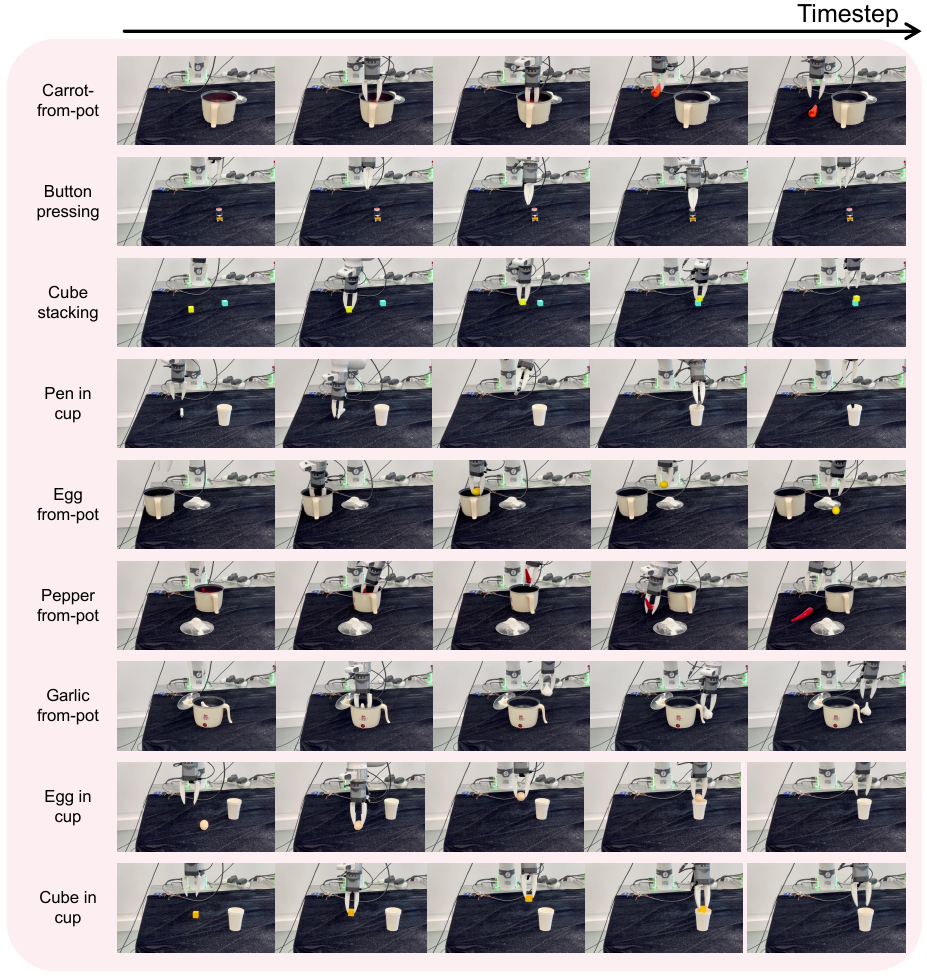}
\caption{\textbf{Detailed visualization for Single arm short range real-world robotic arm task inference demo} We show the inference visualization for Single arm short range real-world task, including In-Domain and Out-of-Domain tasks.}
\label{fig:real_world_infer_task_visual_single_short}
\end{figure}

\mypara{Detailed Visualization of Inference Demos for Each Real-World Robotic Arm Task. }
Fig.~\ref{fig:real_world_infer_task_visual_single_short} shows the inference demos for Single arm short range real-world task, including In-Domain and Out-of-Domain tasks.

\begin{figure}[h]
\centering
\includegraphics[width=0.95\linewidth]{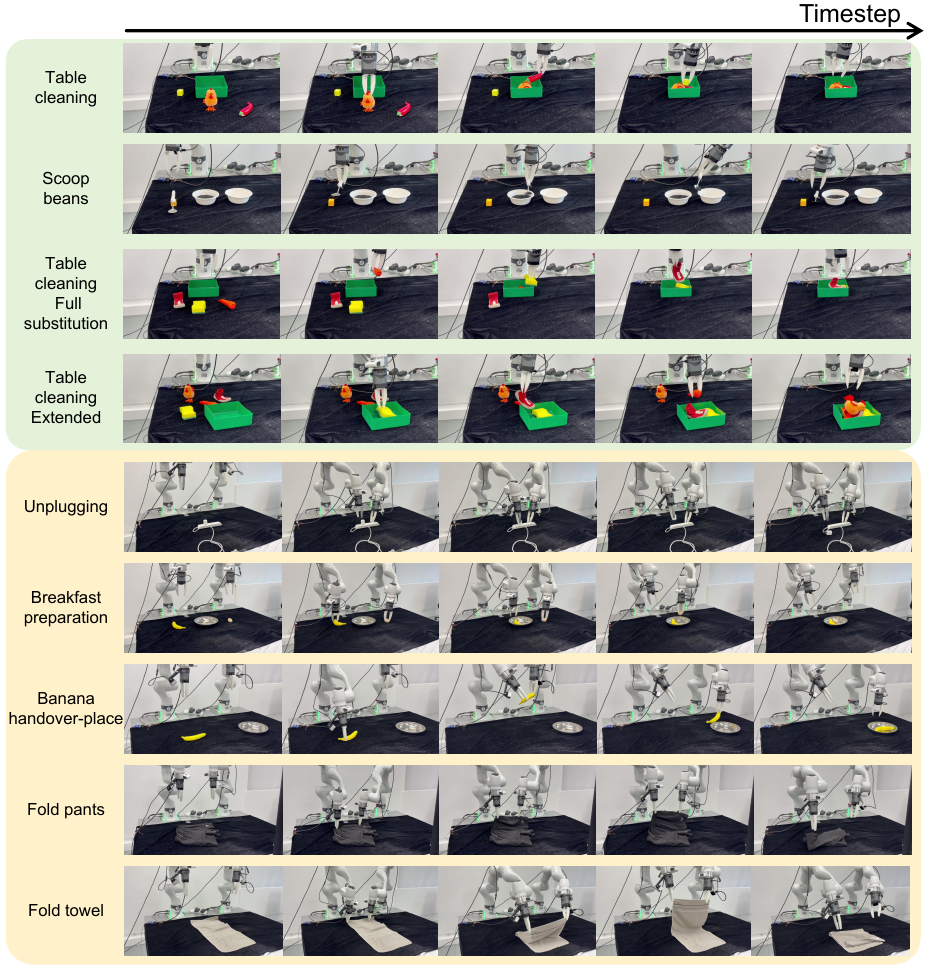}
\caption{\textbf{Detailed visualization for Single arm long horizon and Dual arm coordination real-world robotic arm task inference demo} We show the inference visualization for Single arm long horizon and Dual arm coordination real-world task, including In-Domain and Out-of-Domain tasks.}
\label{fig:real_world_infer_task_visual_single_long_dual}
\end{figure}

\mypara{Detailed Visualization of Inference Demos for Each Real-World Robotic Arm Task.}
Fig.~\ref{fig:real_world_infer_task_visual_single_long_dual} shows the inference demos for Single arm long horizon and Dual arm coordination real-world task, including In-Domain and Out-of-Domain tasks.. 

\section{Future Work and Broader Impact.}

\mypara{Limitations.}
Due to resource constraints, our study focuses on a 4B-scale VLM backbone and does not explore substantially larger models. Larger VLMs may contain stronger visual, spatial, and language priors, and the optimal pre-training recipe may change with model scale. 

\mypara{Broader Impact.}
We hope this work encourages the community to pay more attention to how VLM backbones should be trained for VLA tasks, rather than treating the backbone as a fixed component inherited from general vision-language pre-training. Future work can study how to build VLA-oriented VLMs that could better generalize across tasks, environments, and embodiments. More broadly, improving the training of reusable VLA backbones may make robot learning less dependent on massive proprietary robot datasets, helping more researchers build generalizable VLA models with accessible data and compute.